\documentclass[11pt]{article}

\usepackage[preprint]{acl}

\usepackage{times}
\usepackage{latexsym}

\usepackage[T1]{fontenc}

\usepackage[utf8]{inputenc}

\usepackage{microtype}

\usepackage{inconsolata}

\usepackage{graphicx}

\usepackage{hyperref}       
\usepackage{url}            
\usepackage{booktabs}       
\usepackage{nicefrac}       

\usepackage{pifont}
\usepackage{caption}

\usepackage{fancyvrb}
\usepackage{color}
\usepackage[many]{tcolorbox}
\usepackage{amsmath}
\usepackage{amsfonts}
\definecolor{green_first}{RGB}{168, 209, 176}   
\definecolor{green_second}{RGB}{200, 235, 200}  
\definecolor{green_third}{RGB}{235, 255, 235}   
\definecolor{light_green_table}{RGB}{220, 255, 220}  
\definecolor{light_purple_table}{RGB}{235, 225, 255}  
\definecolor{light_green}{rgb}{0.569, 0.800, 0.459}
\definecolor{blue_dist}{rgb}{0.192,0.443,0.651}
\definecolor{orange_dist}{rgb}{0.812,0.545,0.239}
\definecolor{yellow_dist}{rgb}{0.918,0.804,0.463}
\definecolor{website}{rgb}{0.9333333333333333, 0.10980392156862745, 0.592156862745098} 
\definecolor{pm_rowcolor}{rgb}{0.85, 0.90, 0.84} 
\definecolor{medium_gray}{RGB}{150, 150, 150}  
\definecolor{medium_purple}{RGB}{150, 120, 200}  
\newtcolorbox{promptbox}[2][Prompt]{
    colback=black!5!white,        
    arc=5pt,                      
    boxrule=0.5pt,                
    fonttitle=\bfseries,          
    title=#1,                     
    before upper={\small},        
    fontupper=\fontfamily{ptm}\selectfont, 
    colframe=#2,                  
    left=3pt,                     
    right=3pt,                    
    top=3pt,                      
    bottom=3pt,                   
    boxsep=3pt,                   
    toptitle=1pt,                 
    bottomtitle=1pt,              
    lefttitle=1pt,                
    righttitle=1pt,               
}
\usepackage{fontawesome}
\usepackage{multirow}
\usepackage{subcaption}
\usepackage{wrapfig}
\usepackage{algorithm}
\usepackage{algpseudocode}
\usepackage{amsmath}
\usepackage{float}
\usepackage{verbatim}

\title{SCAN: Structured Capability Assessment and Navigation for LLMs}

\author{%
Zongqi Wang$^{1}$\thanks{{} {} Work done during internship at Baidu, Inc. Contact: \texttt{<zq-wang24@mails.tsinghua.edu.cn>}.}\;\,, \hspace{.3em}\
Tianle Gu$^{1}$, \hspace{.3em}\
Chen Gong$^{2}$, \hspace{.3em}\
Xin Tian$^{3}$, \hspace{.3em}\
Siqi Bao$^{3}$, \hspace{.3em}\
Yujiu Yang$^{1}$\thanks{{} {} Corresponding author: Yujiu Yang \texttt{<yang.yujiu@sz.tsinghua.edu.cn>}.}\;\, \hspace{.3em}\
\\
[1ex]
    $^{1}$ Tsinghua Shenzhen International Graduate School, Tsinghua University \\
    $^{2}$ School of Cyber Engineering, Xidian University \, $^{3}$ Baidu, Inc \\
}

\begin{document}

\maketitle

\begin{abstract}
Evaluating Large Language Models (LLMs) has become increasingly important, with automatic evaluation benchmarks gaining prominence as alternatives to human evaluation. While existing research has focused on approximating model rankings, such benchmarks fail to provide users and developers with a comprehensive and fine-grained understanding of a specific model's capabilities. 
To fill this gap, we propose \textbf{SCAN} (Structured Capability Assessment and Navigation), a practical framework that enables detailed characterization of LLM capabilities through comprehensive and fine-grained evaluation. SCAN incorporates four key components: (1) TaxBuilder, which extracts capability-indicating tags from extensive queries to construct a hierarchical taxonomy automatically; (2) RealMix, a query synthesis and filtering mechanism that ensures sufficient evaluation data for each capability tag; (3) a suite of visualization and analysis tools that facilitate efficient navigation and analysis of model capabilities; and (4) a PC$^2$-based (Pre-Comparison-derived Criteria) LLM-as-a-Judge approach that achieves significantly higher accuracy compared to classic LLM-as-a-Judge method. 
Using SCAN, we conduct a comprehensive evaluation of 21 mainstream LLMs. Our detailed analysis of the GPT-OSS family reveals substantial performance variations, even within sub-capabilities belonging to the same category of capability. This finding highlights the importance of fine-grained evaluation in accurately understanding LLM behavior. 
Project homepage and resources are available at \href{https://github.com/liudan193/SCAN}{https://github.com/liudan193/SCAN}. 
\end{abstract}

\addtocontents{toc}{\protect\setcounter{tocdepth}{0}}

\section{Introduction}

Evaluating large language models (LLMs) has become increasingly important as these models advance in various capabilities~\cite{wang2023pandalm, lin2024wildbench, white2024livebench, li2024llms, li2024c, kim2025biggen, qin2024infobench, liu2024alignbench, gu2024survey, saranathan2024dele, habba2025dove, van2024field, you2024llm, chiang2024chatbot}. Among them, Chatbot Arena~\cite{chiang2024chatbot}, which uses over 2.8 million human-labeled cases, provides relatively accurate evaluations for 223 LLMs (as of 2025-04-09). 
However, its labor-intensive nature limits its applicability, which is particularly pronounced during model development, where timely feedback is essential. 
These limitations have motivated researchers to explore automatic evaluation~\cite{zheng2023judging, dubois2024length, ni2024mixeval, li2024crowdsourced, zhao2024auto, liu2024alignbench}, which uses a small amount of data and LLM-as-a-Judge~\cite{gu2024survey, li2024generation, ashktorab2024aligning, tseng2024expert, son2024llm} to approximate the model rankings in Chatbot Arena. 
These automatic evaluation benchmarks have successfully established a paradigm focused on approximating the rankings of multiple models. 

However, when users and developers aim to gain a comprehensive and detailed characterization of an LLM's capabilities, such a paradigm fails to provide the necessary breadth and granularity. 
To address this critical gap in LLM evaluation, we need an approach that is both comprehensive in scope and fine-grained in detail. 
In this paper, we thoroughly discuss the challenges of constructing a comprehensive and fine-grained evaluation framework and present our solutions, which are integrated into SCAN (Structured Capability Assessment and Navigation), a practical framework that effectively addresses these challenges. 
Specifically, SCAN incorporates four key designs to systematically establish such an evaluation framework.
(1) To enable comprehensive and fine-grained evaluation, we propose TaxBuilder, which extracts capability-indicating tags (e.g., \textit{Python programming}, \textit{physics Q\&A}) from a large volume of real-world human queries. These tags span a broad range of domains, scenarios, and tasks, thereby ensuring wide coverage of diverse user needs.
(2) To facilitate efficient inspection of fine-grained performance, TaxBuilder organizes queries into a hierarchical taxonomy, which is further visualized through a dedicated visualization tool (available on our project page). Together, these components allow users to quickly navigate and analyze model capabilities in a structured and intuitive manner.
(3) The evaluation validity of a specific tag would be unreliable if it is only supported by limited assessment data, which is particularly problematic for tags in the long-tail distribution. To ensure each tag is evaluated with sufficient queries, we develop RealMix, which obtains realistic and high-quality queries through synthesis and filtering. RealMix builds upon TaxBuilder, thereby enhancing both the coverage and diversity. 
(4) Finally, to guarantee evaluation accuracy while maintaining scalability, we propose a novel pointwise LLM-as-a-Judge based on PC$^2$ (Pre-Comparison-derived Criteria). This method significantly improves the reliability and accuracy of automatic judgments while avoiding the prohibitive costs of pairwise evaluation.

To demonstrate the effectiveness of SCAN, we evaluate 21 mainstream LLMs and conduct in-depth analyses of their results. 
A key value proposition of SCAN is its ability to quickly provide comprehensive and detailed understanding of newly released LLMs. To this end, we conducted a detailed analysis of the recently released GPT-OSS family using SCAN. Our analysis reveals several interesting findings:
(1) In terms of overall ranking, GPT-OSS-120B achieves superior performance, surpassing Deepseek-R1, while GPT-OSS-20B shows relatively weak performance, falling below Qwen3-8B.
(2) Despite its overall strong performance, GPT-OSS-120B exhibits limited capabilities in writing and roleplay domains. Conversely, while GPT-OSS-20B generally underperforms, it demonstrates leading capabilities in the coding domain.
(3) Although the GPT-OSS models show strong coding abilities overall, a fine-grained analysis by programming language reveals significant variations: they excel in JavaScript but lag behind in C and Java.
(4) In the knowledge domain, while the GPT-OSS models perform poorly overall, our fine-grained analysis reveals substantial performance variations across different sub-domains, indicating inconsistent coverage of domain-specific knowledge during training.
These fine-grained insights highlight the effectiveness and potential of SCAN.

\noindent In summary, our main contributions are three-fold:

\noindent $\bullet$ We propose shifting evaluation paradigm from rankings to model understanding. To this end, we introduce SCAN, which facilitates comprehensive understanding of LLM capabilities (strengths and weaknesses) through fine-grained evaluation. 

\noindent $\bullet$ We develop the initial version of the taxonomy and evaluation dataset for six domains: writing, roleplay, knowledge, coding, mathematics, and reasoning. These are denoted as SCAN-T-V0 (Fig.~\ref{fig:simple_tree}) and SCAN-D-V0 (Tab.~\ref{tab:SCAN_v0_stat}), respectively. 

\noindent $\bullet$ To showcase the usage of SCAN, we evaluate 21 mainstream LLMs and conduct in-depth analyses of the results. Through detailed analysis of the GPT-OSS family models, we uncover several fine-grained insights, thereby demonstrating the effectiveness and potential of SCAN.


\begin{figure*}[t]
\centering
    \includegraphics[width=\textwidth]{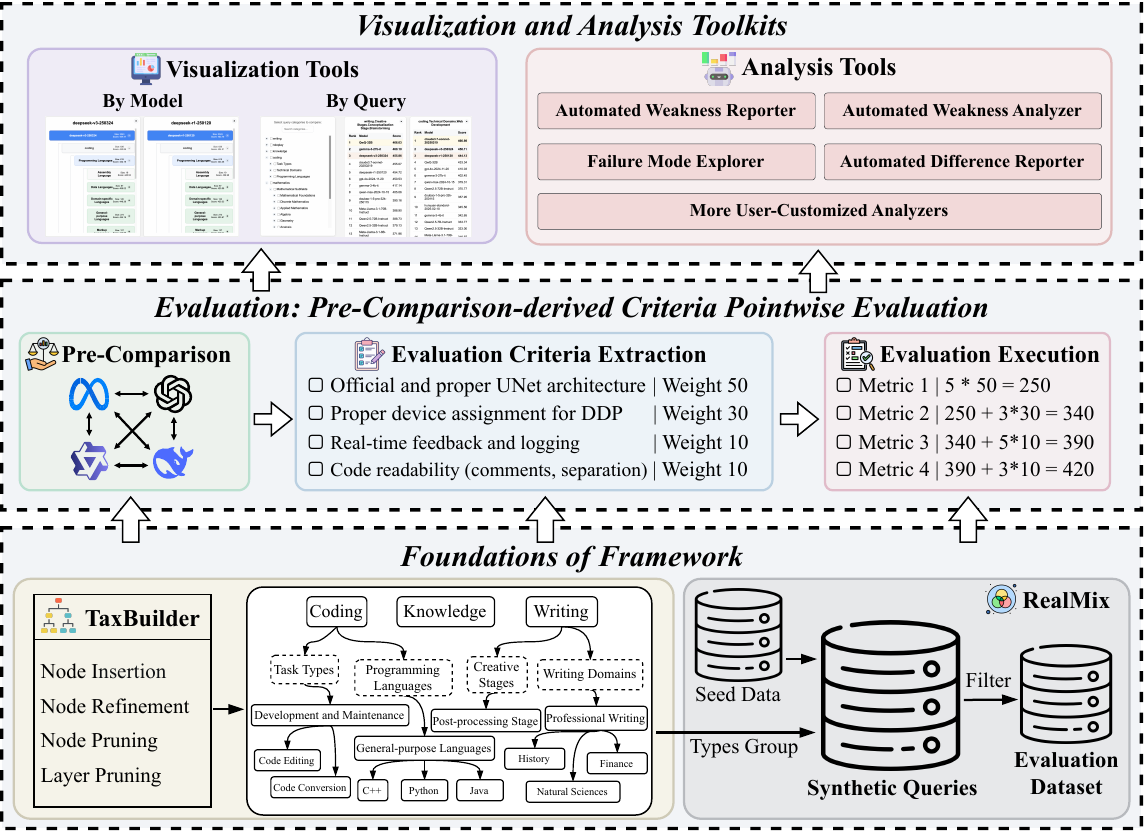}
\caption{An overview of SCAN framework.}
\label{fig:main}
\end{figure*}

\section{SCAN}
\label{sec:dataset_section}

The overview of SCAN is shown in Fig.~\ref{fig:main}. It consists of three components: TaxBuilder for building an extensible query taxonomy, RealMix for generating new queries based on the taxonomy, and visualization and analysis toolkits for user-friendly interaction. 

\subsection{TaxBuilder}
\label{sec:automatic_categorization}


\begin{figure}[t]
\centering
    \includegraphics[width=0.475\textwidth]{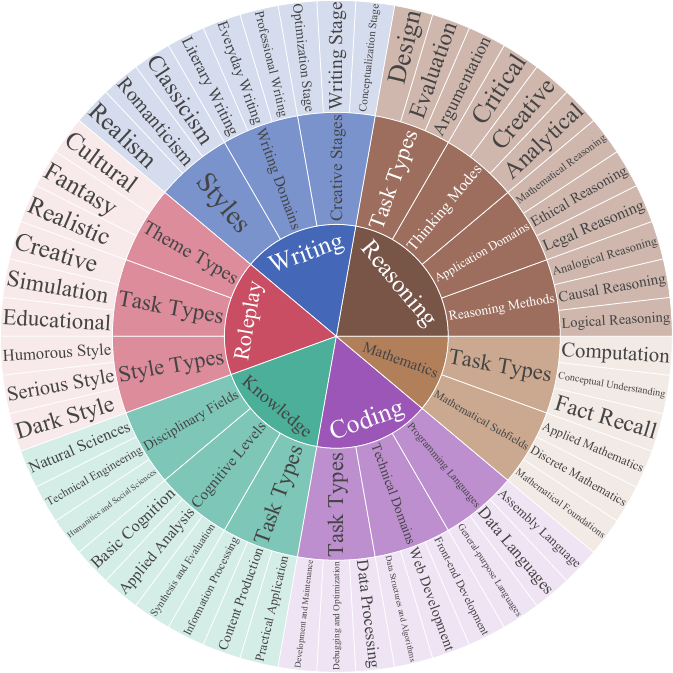}
\caption{A subset of SCAN-T-V0. Please refer to project page for complete taxonomy.}
\label{fig:simple_tree}
\end{figure}

\begin{figure*}[t]
    \centering
    \includegraphics[width=\textwidth]{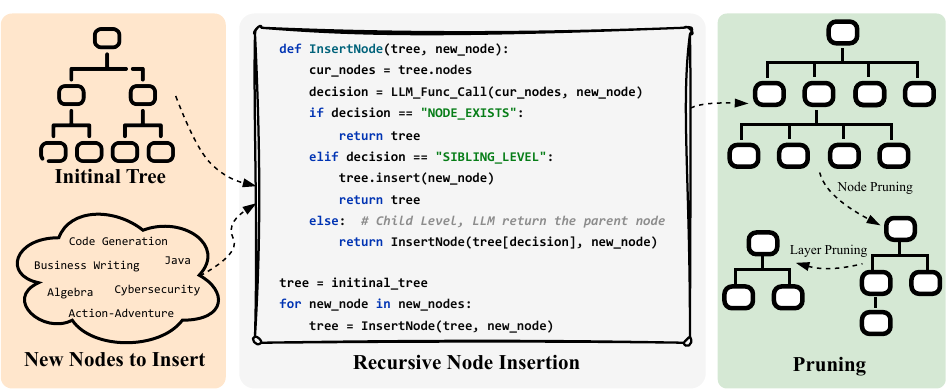}
    \caption{An overview of TaxBuilder, a tree-based automatic taxonomy generation.}
    \label{fig:auto_categorization}
\end{figure*}

To establish a comprehensive and fine-grained evaluation system, we first need to develop the query taxonomy. To achieve this, we introduce TaxBuilder, an automatic and extensible solution for constructing tree-based taxonomy from large volumes of unstructured queries. The workflow of TaxBuilder is illustrated in Fig.~\ref{fig:auto_categorization}. 

\noindent\textbf{Preparation.} TaxBuilder initiates the process with a manually constructed basic taxonomy tree $T_{init}$. This basic tree is simple and requires minimal human effort (refer to \S~\ref{sec:appendix_TaxBuilder_basic_tree} for details). 
Then, we employ a low-cost model (gpt-4o~\cite{openai2024gpt4o}) to annotate query tags for a large set of real user queries \footnote{We utilize the dataset made available by the Chatbot Arena team, accessible at \url{https://www.kaggle.com/competitions/wsdm-cup-multilingual-chatbot-arena/data}.}. 
The primary aim of this phase is to generate a large pool of unstructured query tags, named $TG_{init}$. 
Further details regarding this step are elaborated in \S~\ref{sec:appendix_TaxBuilder_init_tags}. 

\noindent\textbf{Node Insertion.}
The most core design of TaxBuilder is its node insertion mechanism. A naive approach would be to input the entire taxonomy as context to a powerful LLM and ask it to identify the position for the new node. However, this task involves both long-context and complex reasoning, which is challenging even for the most advanced LLMs. Moreover, as the taxonomy grows, the input context becomes increasingly longer and the reasoning more difficult, thereby limiting the extensibility of this approach. 

To address this issue, we introduce a recursive node insertion process inspired by how trees are constructed in data structures. When a new node needs to be inserted, we traverse the tree step by step. At each step, the LLM only needs to determine the relationship between the new node and the nodes at the current level. In this way, the model only considers a small portion of the tree and reasons over three options: (1) the node already exists, (2) it is a sibling of the current nodes, or (3) it should be added as a child of one of them. This design keeps the context short and the reasoning simple, making TaxBuilder significantly more scalable as the taxonomy expands. 


Formally, given a node \( tg \) and the current tree \( T_{cur} \) (we represent the tree by its root), we invoke LLM-as-Decision-Maker, denoted as \( DM_{ins} \), as follows:

\begin{equation}
    DM_{ins}(tg, T^{c}_{cur}) \rightarrow \{ \text{<E>}, \text{<S>}, T^{c_i}_{cur} \},
\label{eq:dm_ins}
\end{equation}

\noindent where \( T^{c}_{cur} = \{ T^{c_1}_{cur}, T^{c_2}_{cur}, \ldots, T^{c_m}_{cur} \} \) denotes the set of first-level child nodes of \( T_{cur} \). The function of \( DM_{ins} \) is to determine the relationship between \( tg \) and \( T^{c}_{cur} \). The prompt for \( DM_{ins} \) is presented in Fig.~\ref{fig:dm_ins}. It will return one of three possible relationship types:

\begin{itemize}
    \item \textbf{<E> (Exists):} This indicates that the node \( tg \) already exists in \( T^{c}_{cur} \) or has a semantically equivalent node. In this case, the new node \( n \) is discarded to avoid redundancy. 
    
    \item \textbf{<S> (Sibling):} This signifies that \( tg \) is a sibling of the nodes \( T^{c}_{cur} \). Consequently, \( T^{c}_{cur} \) is updated to include \( tg \), i.e., \( T^{c}_{cur} = \{ T^{c_1}_{cur}, T^{c_2}_{cur}, \ldots, T^{c_m}_{cur} \} \cup \{ tg \} \). 

    \item \textbf{\( \mathbf{T^{c_i}_{cur}} \) (Child Node):} This implies that \( tg \) should be a child of the specific node \( T^{c_i}_{cur} \). In this scenario, a recursive insertion is performed. 
\end{itemize}

After recursively traversing the entire tree, we successfully insert the node \( tg \) and obtain the tree for the next iteration. 
We iteratively perform this operation for all nodes, and ultimately obtain a tree that contains all tags from \( TG_{init} \). 


\noindent\textbf{Node Refinement and Pruning.} 
Node refinement serves two purposes: (1) resolving incorrect parent-child relationships between nodes at the same level, and (2) splitting leaf nodes that conflate multiple concepts into separate, more atomic concepts (e.g., separating "science fiction fanfiction" into "science fiction" and "fanfiction"). 
Node pruning addresses two issues: (1) merging duplicate nodes at the same level, and (2) removing rarely encountered concepts that unnecessarily complicate the tree. 
We perform the refinement and pruning processes recursively, guided by the LLM-as-Decision-Maker (see Fig.~\ref{fig:dm_node_refine_prun}), which effectively corrects faulty nodes and controls node complexity. 

\noindent\textbf{Layer Pruning.} 
We introduce two rules to reduce tree depth: (1) Meaningless depth pruning: if a parent node has only a single child, the child is pruned to eliminate unnecessary depth; (2) Complexity pruning: the tree is restricted to a maximum of 4 layers. Nodes beyond the 4 layer are retained only if they have 5 or more child nodes, as such structures are usually meaningful. which effectively controls tree complexity and prevents overgrowth. 

After completing node insertion, node refinement, node pruning, and layer pruning, we obtain the final taxonomy, which we call SCAN-T-V0. The current taxonomy tree is easy to interpret by human, allowing for convenient manual modification if needed. A simplified version of this tree is shown in Fig.~\ref{fig:simple_tree}. The formal algorithm for the tree construction process is provided in Alg.~\ref{alg:tax_builder}. 

\subsection{RealMix}
\label{sec:query_generation}

We opt for synthetic data over existing datasets for two reasons: (1) existing datasets may not contain enough queries with the required tags in taxonomy, and (2) to prevent data contamination. 
To ensure the evaluation data includes high-quality, realistic queries that can be encountered in real-world, we propose RealMix. RealMix operates by mixing several appropriate content details from multiple real-user queries and adapting them to align with the specified tags. 
While there are other data synthesis approaches~\cite{liu2024regmix, liu2025quadmix} that mix multiple queries, none are capable of generating queries with specified tags. The core advantage of RealMix in fact lies in that it is built upon the TaxBuilder, which enables controlled and tag-aligned synthesis. 

\noindent\textbf{Seed Data.} We first use QwQ-32b~\cite{qwq32b, qwen25} to annotate a set of real-user queries~\cite{wsdm_cup_multilingual_chatbot_arena} with three types of labels: domain, query tags, and quality. The domain label covers six domains: writing, roleplay, knowledge, coding, mathematics, and reasoning. Query tags are assigned based on SCAN-T-V0. For quality annotation, each domain is paired with a checklist of seven criteria; queries meeting at least four are marked as high-quality. In total, we collect approximately 31,000 high-quality annotated queries. Further details are provided in \S~\ref{sec:appendix_realmix_seed_data}. 

\noindent\textbf{Query Synthesis.}
RealMix leverages content from real user queries to generate queries aligned with specific tags. This approach allows us to obtain realistic queries associated with desired tags. To synthesize a query, we sample one reference query and three content queries. A query generation model then extracts appropriate real-world content from the content queries and generates a new query that incorporates these contents while matching the tags of the reference query. If no suitable content is found for the specified tags, this generation is discarded. 
We adopt reference queries instead of random tag combinations from SCAN-T-V0 since some combinations do not exist in practice. To mitigate model bias, we employ multiple query generation models (Deepseek-V3~\cite{deepseekai2024deepseekv3technicalreport}, gpt-4o, and doubao-1.5pro~\cite{doubao2024pro}). More details are provided in \S~\ref{sec:appendix_realmix_syn_query}. 

\noindent\textbf{Verification and Filtration.}
We validate both the tags and the quality of generated query using the same procedure applied to the seed data, and discard low-quality queries. To further reduce model bias, for each query, we randomly sample checking models for quality and tags from three reasoning models (QwQ-32B, Deepseek-R1~\cite{deepseekai2025deepseekr1incentivizingreasoningcapability}, and o1-mini~\cite{openai2024o1mini}). 

\subsection{Visualization and Analysis Toolkits} 
\label{sec:vis_tool} 

We provide a suite of visualization and analysis tools to assist users in understanding and interpreting evaluation results. Due to space constraints, the description of each tool is provided in \S~\ref{sec:vis_and_analysis_tool}. 



\section{Dataset Description}
\label{sec:dataset}

\begin{figure}[!h]
    \centering
    \begin{minipage}{0.5\textwidth}
        \centering
        \resizebox{0.93\textwidth}{!}{
        \begin{tabular}{lcccc}
        \toprule
            \textbf{Domain} & \textbf{\#Samples} & \textbf{\#Tags} & \textbf{\#MinNum} & \textbf{Avg.Length} \\
            \midrule
            Writing     & 1108  & 594 & 19  & 772.57 \\
            Roleplay    & 470   & 429 & 19  & 1008.46 \\
            Knowledge   & 540   & 315 & 20  & 608.57 \\
            Coding      & 636   & 369 & 19  & 1232.03 \\
            Mathematics & 344   & 189 & 20  & 817.02 \\
            Reasoning   & 245   & 186 & 19  & 904.81 \\
            \rowcolor{light_green_table} Total  & 3343  & 2082 & 19  & 880.92 \\
        \bottomrule
        \end{tabular}}
        \captionof{table}{Statistics of SCAN-D-V0.}
        \label{tab:SCAN_v0_stat}
    \end{minipage}%
    \hfill
    \begin{minipage}{0.45\textwidth}
        \centering
        \includegraphics[width=\textwidth]{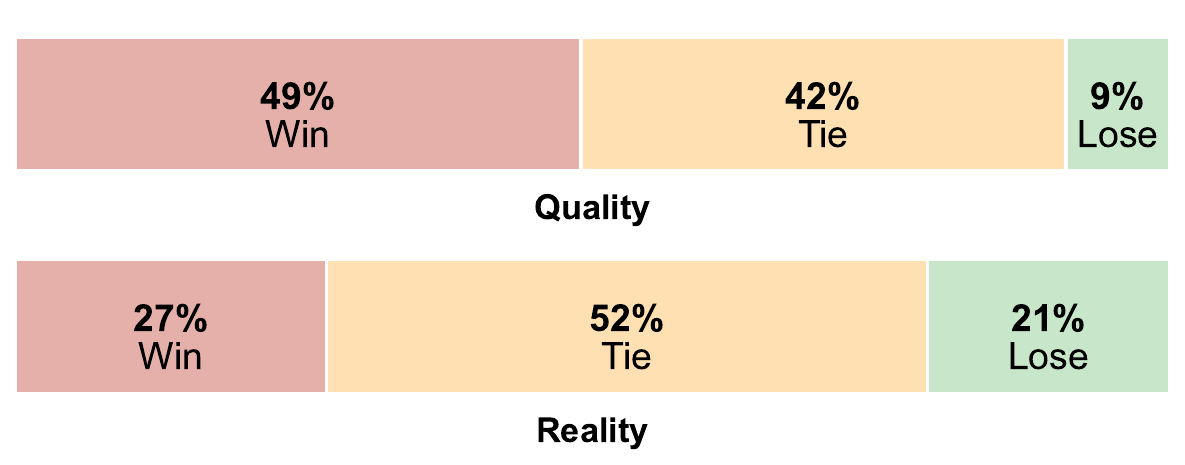}
        \caption{Comparison of RealMix and real user queries, judged by five human evaluators.}
        \label{fig:quality_human_eval}
    \end{minipage}
\end{figure}

\noindent\textbf{Dataset Statistics.} 
We construct SCAN-D-V0, the initial version of the evaluation dataset, which consists of 3,343 queries, each annotated with a domain and tags. 
Tab.~\ref{tab:SCAN_v0_stat} summarizes the key statistics of SCAN-D-V0. 
We ensure that the sample size for each tag in SCAN-T-V0 is at least 19 (\#MinNum) to guarantee the statistical representativeness of the evaluation results. 

%

\noindent\textbf{Quality Control.} We validate data quality through three tasks involving human and automatic evaluation: Five graduate-level annotators perform two comparative assessments. For each generated query and its corresponding real user query (i.e., reference query), annotators evaluate (1) which query is higher in quality and (2) which is more likely to occur in real-world scenarios. Results in Fig.~\ref{fig:quality_human_eval} show generated queries are consistently preferred, confirming the effectiveness of RealMix. Details are in \S~\ref{sec:appendix_evaluation_dataset_human_quality}. 
(3) We assess potential data leakage in generated queries in \S~\ref{sec:appendix_evaluation_dataset_data_contamination}. 


\section{Pre-Comparison-derived Criteria-based LLM-as-a-Judge}
\label{sec:evaluate_method}

In this section, we first formulate traditional evaluation paradigms and highlight their strengths and limitations in Sec.~\ref{sec:problem_formulation}. 
Then, we present the motivation and description of our method in Sec.~\ref{sec:our_method}. 

\subsection{Problem Formulation}
\label{sec:problem_formulation}







\noindent\textbf{Naive Pointwise Evaluation.} Naive pointwise evaluation~\cite{kim2023prometheus, ye2024self} refers to scoring each response independently. Judge model $J$ is tasked to assign a score to a single response $y$ with a naive pointwise evaluation prompt $p_{o}$. 

\begin{equation}
J(y|x,p_{o}) \rightarrow s \in \mathbb{R}.
\end{equation}

Pointwise evaluation is straightforward and does not incur unacceptable evaluation costs as the number of models being evaluated increases. However, it often produces inferior evaluation results~\cite{li2024llms, zheng2023judging}. 

\noindent\textbf{Pairwise Evaluation.} Pairwise evaluation~\cite{cao2024compassjudger, hu2024rethinking} compares two responses to determine which is better. According to prior studies~\cite{zheng2023judging, li2024llms}, while this approach typically produces more reliable results than pointwise evaluation, it has a quadratic time complexity O(n$^2$) that becomes impractical as the number of tested models increases. Therefore, we do not adopt it in our work. 

\subsection{Methodology}
\label{sec:our_method}

\noindent\textbf{Motivation.} 
Our core insight is that, given the superior performance of pairwise evaluation over pointwise evaluation, we can adapt the mechanism of pairwise evaluation into the pointwise setting to create an enhanced pointwise evaluation. 
Pairwise evaluation excels in comparing two responses by identifying their differences, evaluating which response performs better with respect to each difference, and ultimately selecting the winner based on advantages in the most important differences. (See examples in \S~\ref{sec:supplementary_material_of_eval_method}.)
In contrast, pointwise evaluation struggles to identify such differences because it evaluates each response independently. 
To bridge this gap, we introduce a pre-comparison phase, where we leverage several auxiliary responses to extract relevant evaluation criteria before scoring. These criteria act as proxies for the key differences identified in pairwise evaluation. At the same time, we prompt the model to assign weights to each criterion, emulating how pairwise evaluation prioritizes more significant differences. 

\noindent\textbf{Pre-Comparison-derived Criteria Extraction.} 
The core design of PC$^{2}$ pointwise evaluation lies in extracting criteria from a pre-comparison process. 
Specifically, given an instruction $x$, we prompt $n$ LLMs $\{M_1, M_2, ..., M_n\}$ to produce $n$ responses $\{y^1, y^2, ..., y^n\}$. The selection of multiple LLMs is intended to diversify the responses, thereby facilitating the extraction of comprehensive, diverse and effective criteria. 
Subsequently, we assign the judge model $J$ two tasks. The first task involves using $J$ to compare the $n$ responses, thereby surfacing their differences (i.e., criteria). This is done using prompt $p^c$. The second task is to weight these criteria using prompt $p_w$: 


\begin{flalign}
&J(\{y^1, y^2, \dots, y^n\} \mid x, p_c, p_w) \rightarrow & \notag \\[0.3em]
&\quad C = \{c_1, c_2, \dots, c_m\},\;
W = \{w_1, w_2, \dots, w_m\} && 
\end{flalign}

\noindent where $C$ represents the criteria, and $W$ denotes the corresponding weights, where $\sum_{i=1}^{m} w_i = 100$. Since we use a reasoning LLM as the judge model, we typically assign two tasks to the model simultaneously. If the judge model is a chat model, we recommend handling these tasks separately. 

\noindent\textbf{Evaluation Execution.} Once the criteria are determined, we can score a single response $y$:

\begin{equation}
J(y|x,p_c,y_b,C,W) \rightarrow s \in \mathbb{R},
\end{equation}

\noindent where $p_c$ represents our criteria-weighted evaluation prompt and $s$ is the final score. For the baseline answer $y_b$, we first evaluate it without using any baseline answer, then concatenate the evaluation results with $p_c$ to assess other models. This operation helps improve the stability of multiple evaluations. 

This approach leverages the comprehensive understanding gained through pre-comparison to establish query-specific criteria, enabling more effective evaluation while maintaining computational efficiency as the number of evaluated models increases. 

\section{Results on PC\texorpdfstring{$^{2}$}{2} Pointwise Evaluation}
\label{sec:exp_method}

In this section, we demonstrate the main experimental results. The experiment details are shown in \S~\ref{sec:supplementary_material_of_experiment_evaluation_method}. Results on more benchmark and ablation results are shown in \S~\ref{sec:appendix_difference}. 


\begin{table}[t]
\centering
\resizebox{0.45\textwidth}{!}{
    \begin{tabular}{lc}
    \toprule
    \textbf{Method / Model} & \textbf{Accuracy} \\
    \midrule
    \multicolumn{2}{l}{\textbf{\texttt{Deepseek-R1}}} \\
    \quad naive & 0.5694 \\
    \quad direct metric decomposition & 0.6134 \\
    \quad metric decomposition (single model) & 0.5974 \\
    \quad metric decomposition (diverse model) & 0.6466 \\
    \quad ours & \textbf{0.6962} \\
    \midrule
    \multicolumn{2}{l}{\textbf{\texttt{Qwen3-32B}}} \\
    \quad naive & 0.5181 \\
    \quad ours & \textbf{0.6535} \\
    \midrule
    \multicolumn{2}{l}{\textbf{\texttt{claude-3.7-sonnet}}} \\
    \quad naive & 0.5959 \\
    \quad ours & \textbf{0.7453} \\
    \midrule
    \multicolumn{2}{l}{\textbf{\texttt{gpt-4.1}}} \\
    \quad naive & 0.6116 \\
    \quad ours & \textbf{0.7201} \\
    \bottomrule
    \end{tabular}}
\caption{Comparison of accuracy across different LLM-as-a-Judge methods.}
\label{tab:evaluation_method_performance_acl2026}
\end{table}

As shown in Tab.~\ref{tab:evaluation_method_performance_acl2026}, naive pointwise evaluation yields relatively low accuracy across all tested LLMs, confirming its limited reliability as a judging approach. Introducing metric decomposition markedly improves performance, with direct decomposition and single-model pre-comparison offering moderate gains, and diverse-model pre-comparison delivering more substantial improvements. Building upon this, our method further incorporates baseline answers, leading to consistent and significant accuracy gains over all baselines. These results demonstrate that the combination of criteria decomposition, diverse-model pre-comparison, and baseline guidance forms a robust and generalizable evaluation strategy.

\section{Results on SCAN}
\label{sec:exp_SCAN}

\subsection{Overview of Evaluation Results}
\label{sec:benchmark}

To utilize SCAN for analysis, we first employ SCAN-D-V0 to evaluate 21 mainstream LLMs in this section. The experimental setup is provided in \S~\ref{sec:appendix_exp_setup_leaderboard}. An overview of evaluation results is presented in \S~\ref{sec:leaderboard}. For complete results, please refer to the project homepage. 

\subsection{Exploring SCAN: GPT-OSS as Examples}
\label{sec:fine_grained_results_gpt_oss}

Upon the release of a new model, our SCAN framework facilitates a rapid, comprehensive evaluation of its capabilities across diverse domains. Here, we demonstrate SCAN's utility through an analysis of the recently released GPT-OSS family models. 

\noindent\textbf{Overview.} The overall evaluation results summarized in Tab.~\ref{tab:leaderboard} show that GPT-OSS-120B achieves the highest aggregate score, reflecting its strong overall performance. However, it ranks only 7th in roleplay domain, highlighting a notable weakness obscured by the composite metric. 
Crucially, our framework reveals that GPT-OSS-20B, despite a lower overall rank, performs exceptionally in coding domain (ranking 2nd), which demonstrates its specialized strength. This underscores the effectiveness of our approach in uncovering fine-grained capability profiles and identifying performance that holistic scores alone cannot capture. 

\begin{figure}[t]
\centering
    \includegraphics[width=0.45\textwidth]{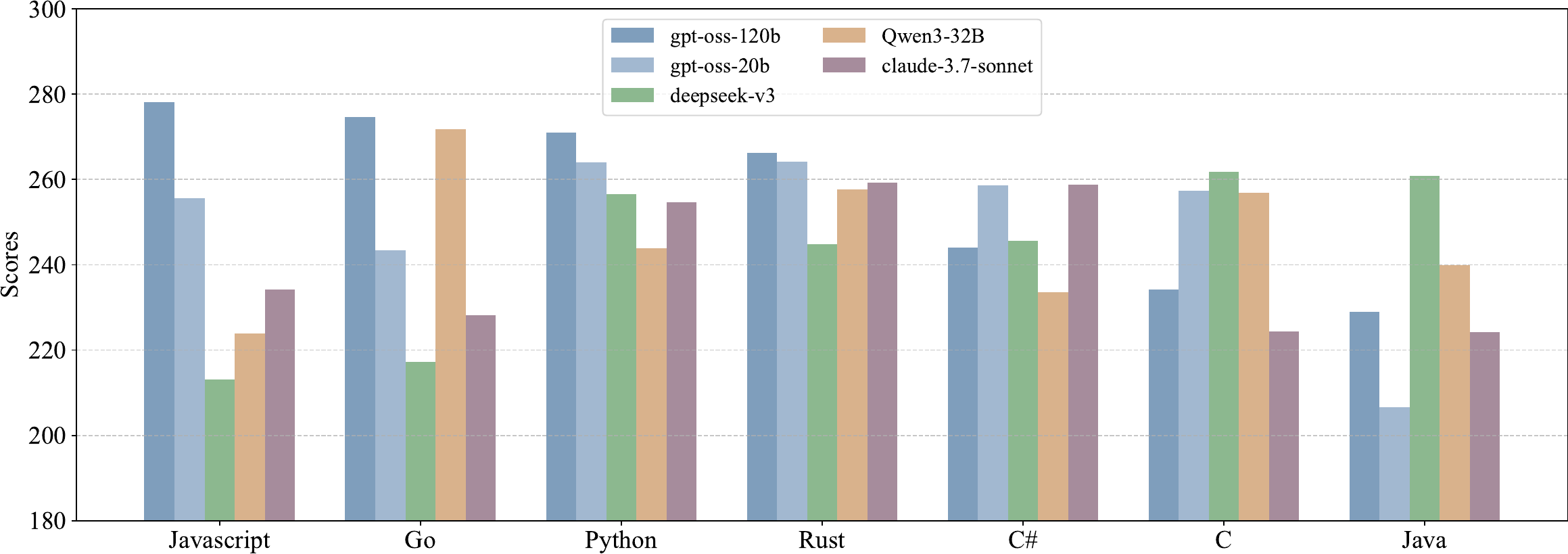}
\caption{Fine-grained performance comparison on \texttt{coding.Programming Languages.General-purpose Languages}.}
\label{fig:fine_grained_performance_on_general_purpose_languages}
\end{figure}

\begin{figure*}[t]
\centering
    \includegraphics[width=0.90\textwidth]{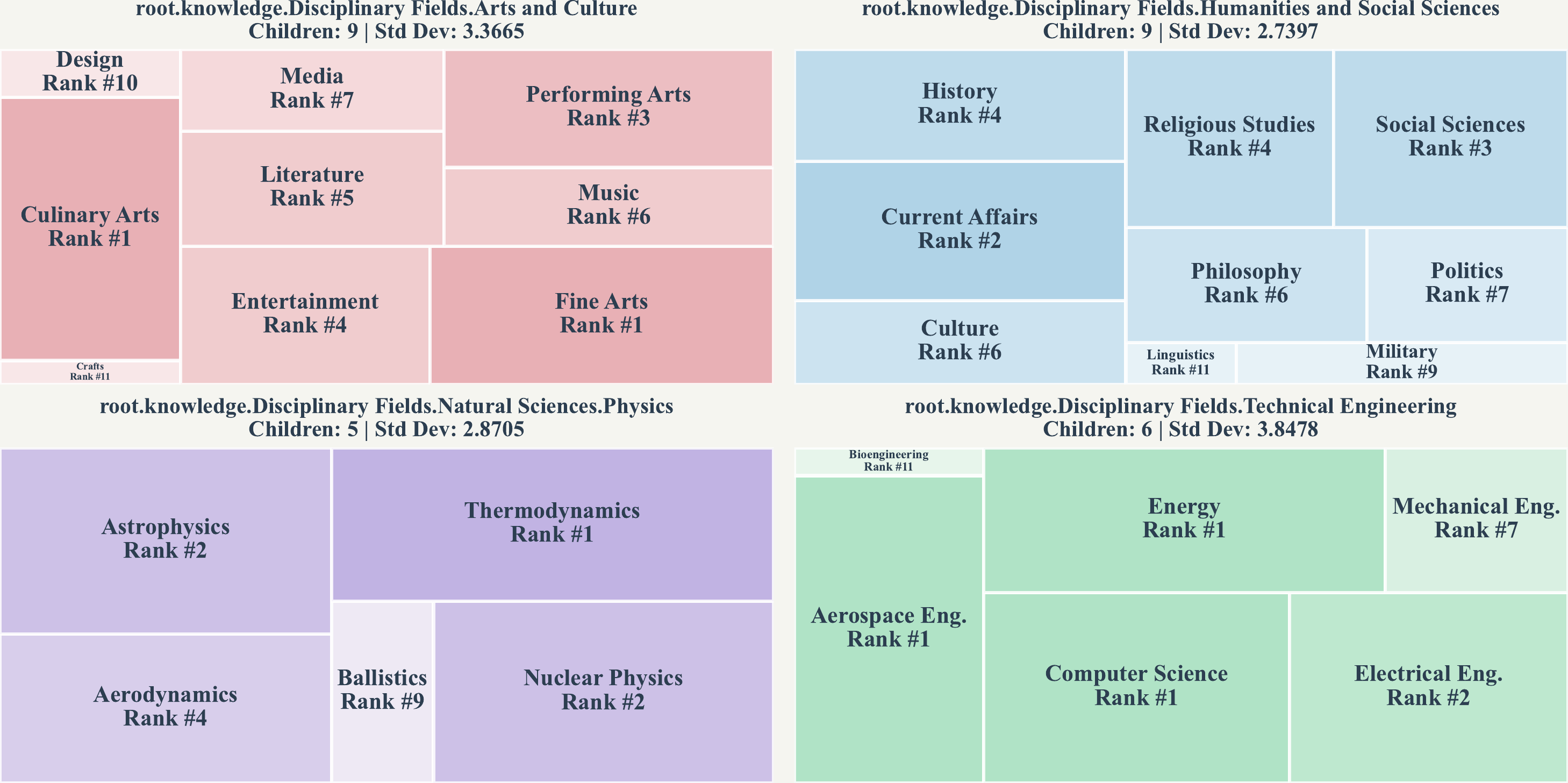}
\caption{The top 4 parent nodes exhibiting the largest performance variance across child nodes, automatically identified by our Failure Mode Explorer tool. The symbol `\#' denotes rankings.}
\label{fig:fine_grained_performance_on_knowledge}
\end{figure*}

\noindent \textbf{Fine-grained Coding Performance Analysis.} Our SCAN framework enables decomposition of the aggregate coding score by programming language, allowing evaluation of model performance across specific general-purpose languages such as C and Java. The results shown in Fig.~\ref{fig:fine_grained_performance_on_general_purpose_languages} reveal substantial performance variation across languages. 
Notably, GPT-OSS-120B performs strongly in Python, JavaScript, Go, and Rust, but shows weaker performance in C and Java—areas where Deepseek-V3 achieves superior results. This level of granularity is essential for applications demanding language-specific usage. 
Furthermore, the analysis uncovers the "spiky" capability profile of GPT-OSS-20B: it ranks 2nd in Python and Rust, and outperforms its larger counterpart, GPT-OSS-120B, in both C and C\#. This demonstrates that smaller models can excel in specific tasks, providing a more efficient option for domain-specific development. 
SCAN’s fine granularity thus facilitates informed, use-case-driven model selection, advancing beyond holistic rankings toward nuanced capability profiling. 

\noindent \textbf{Diagnosing Knowledge Instability through Fine-Grained Analysis.} 
Here, we demonstrate the use of our analytical tool, Failure Mode Explorer (see \S~\ref{sec:vis_and_analysis_tool}). In Fig.~\ref{fig:fine_grained_performance_on_knowledge}, we present the tool's automatically identified 4 parent nodes with the largest performance variance across sub-nodes for GPT-OSS-120B. 
Interestingly, all four are within the knowledge domain. Combined with the observation from Tab.~\ref{tab:leaderboard} that GPT-OSS-120B's overall performance in knowledge is unexceptional, this motivates a deeper investigation into its capabilities within this domain. 

While GPT-OSS-120B ranks third overall in the knowledge domain, a closer look reveals that its performance is far from uniformly strong. In GPT-OSS-120B’s case, the internal performance distribution is highly uneven. For example, within the technical engineering field (parent node ranked \#2), the model achieves top positions in computer science and aerospace engineering (\#1), yet falls sharply to \#11 in bioengineering. 
This demonstrates that, while aggregate scores facilitate macro-level comparisons, they may conceal important weaknesses in a model’s capabilities. 

To tackle these hidden weaknesses, our SCAN framework converts coarse-grained aggregate metrics into fine-grained diagnostic signals that expose exactly where the gaps lie. In the case above, instead of settling for a general verdict like “the model’s knowledge capability is weak,” SCAN can pinpoint the specific sub-domains—such as bioengineering—that are underperforming relative to other strengths. 
It is worth noting that these results were automatically detected by our Failure Mode Explorer tool, further validating SCAN’s effectiveness in systematically revealing hidden problems. 

\section{Related Work}

\textbf{Automatic Evaluation Benchmark. }
We review existing benchmarks here, focusing on their motivations, dataset designs, and evaluation methods. 
AlpacaEval-2.0~\cite{dubois2024length} uses 805 curated general-domain data points to assess instruction-following with win rate metric. 
MT-Bench~\cite{zheng2023judging} includes 80 high-quality queries across 8 domains. It adopts pointwise evaluation on a 1–10 scale. 
Arena-Hard~\cite{li2024crowdsourced} filters 500 challenging Chatbot Arena datasets and uses win rate metric for evaluation. 
LiveBench~\cite{white2024livebench} manually updates queries in real-time with rule-based scoring to reduce LLM biases. 
Auto-Arena~\cite{zhao2024auto} employs pairwise debates for Elo-based ranking, with a 40-query evaluation set. 
WildBench~\cite{lin2024wildbench} updates real user queries to reflect real-world distributions, using both win rate and pointwise evaluation. 
We discuss the differences between our methods and existing studies in \S~\ref{sec:appendix_difference}. 

\section{Conclusion}

In this paper, we propose shifting the focus of evaluation from only rankings and leaderboards to providing valuable insights. 
To this end, we introduce SCAN, an extensible evaluation framework featuring comprehensive and fine-grained queries, along with tools for visualization and analysis. 
Additionally, our proposed PC$^{2}$-based LLM-as-a-Judge achieves superior accuracy compared to classic evaluation methods. 
Finally, through extensive experiments, we demonstrate that utilizing SCAN enables the convenient identification of a model's strengths and weaknesses, thereby providing valuable insights.

In the future, we believe SCAN will provide valuable insights to the community in the following aspects.
(1) For model evaluation, SCAN helps prevent one-sided understanding of a model’s capabilities, as its broad coverage across diverse categories. 
(2) For model training, SCAN can help identify a model’s strengths and weaknesses; by quantifying fine-grained performance, it enables developers to adjust training parameters accordingly, such as data distribution.
(3) For data synthesis, TaxBuilder offers a scalable taxonomy that empowers developers to generate comprehensive, diverse, and complex synthetic data for model training. 

\section*{Limitations}

Currently, our framework is designed primarily for language models, and does not yet provide support for multimodal models~\cite{li2024survey, patraucean2023perception, li2024seed, roberts2025zerobench, liang2021multibench}. In future work, we aim to extend the framework to include multimodal capabilities through a dedicated module, tentatively named SCAN-Anything. 

In addition, the current implementation does not encompass evaluation dimensions related to AI mechanisms such as safety, honesty, factuality, hallucination, and fairness~\cite{ren2024safetywashing, rottger2025safetyprompts, gu2024mllmguard, jin2024fairmedfm, wang2024ceb, liu2025thinking, luo2024halludial, ravi2024lynx, ding2024hallu, huang2023flames}. Nevertheless, the framework is inherently extensible and can be readily adapted to incorporate these aspects as additional evaluation domains in subsequent work. 



\bibliography{custom}

\newpage
\addtocontents{toc}{\protect\setcounter{tocdepth}{10}}
\appendix
\onecolumn
\renewcommand{\contentsname}{Appendices Contents}
\tableofcontents

\newpage
\twocolumn

\newcommand{\smallerfont}{\fontsize{5}{6}\selectfont}
\begin{table*}[!t]
\centering
\smallerfont
\resizebox{\textwidth}{!}{ 
\begin{tabular}{lcccccc}
    \toprule
    \textbf{Dataset}    & \textbf{\#Queries} & \textbf{Cont.-free} & \textbf{Fin.Tag}  & \textbf{LongTail} & \textbf{Update}   \\
    \midrule
    \textbf{AlpacaEval2.0}       & 805      & \ding{53} & \ding{53} & \ding{53} & \ding{53} \\
    \textbf{MT-Bench}            & 80       & \ding{53} & \ding{53} & \ding{53} & \ding{53} \\
    \textbf{Arena-Hard}           & 500      & \ding{53} & \ding{53} & \ding{53} & \ding{53} \\
    \textbf{Auto-Arena}          & 40       & \ding{51} & \ding{53} & \ding{53} & \ding{51} \\
    \textbf{LiveBench(2024-11-25)} & 300    & \ding{51} & \ding{53} & \ding{53} & \ding{51} \\
    \textbf{MixEval}             & 4000    & \ding{53} & \ding{53} & \ding{53} & \ding{53} \\
    \textbf{MixEval-Hard}        & 1000     & \ding{51} & \ding{53} & \ding{53} & \ding{53} \\
    \textbf{WildBench}           & 1024     & \ding{53} & \ding{53} & \ding{53} & \ding{53} \\
    \textbf{C$^2$LEVA}           & 16115   & \ding{51} & \ding{51} & \ding{53} & \ding{53} \\
    \rowcolor{light_green_table} \textbf{Ours}                & 3343  & \ding{51} & \ding{51} & \ding{51} & \ding{51} \\
\bottomrule
\end{tabular}}
\caption{Comparison of LLM alignment benchmark datasets. “Real.” denotes real-world queries, “Cont.-free” means contamination-free, “Fin.Tag” indicates fine-grained tagging.}
\label{tab:dataset_comparison}
\end{table*}

\section{More Discussion}
\label{sec:more_discussion}



\subsection{Distinctions from Prior Works Enabled by Fine-Grained Dataset}
\label{sec:appendix_difference}

Several existing studies, such as SELF-INSTRUCT~\cite{wang2022self}, Nemotron-4-340B~\cite{adler2024nemotron}, WildBench~\cite{lin2024wildbench}, CLEVA~\cite{li2023cleva}, and C$^{2}$LEVA~\cite{li2024c}, provide fine-grained evaluation of LLM capabilities. However, their taxonomies are relatively coarse and lack extensibility. For instance, C$^{2}$LEVA classifies LLM abilities into four broad categories—Language, Knowledge, Reasoning, and Harms—and further subdivides Language into only two types: Typo-Fixing and Transliteration. In contrast, our benchmark introduces an unprecedentedly detailed, fine-grained, and extensible taxonomy. Additionally, none of the aforementioned benchmarks provide visualization or analytical tools tailored for fine-grained evaluation, which is a key contribution of our work. 

A concurrent study, EvalTree~\cite{zeng2025evaltree}, also organizes queries into a tree structure. However, our TaxBuilder differs fundamentally. 
(1) Unlike EvalTree, which reorganizes existing datasets, our method starts with a fixed, extensible taxonomy and automatically synthesizes data to populate it. 
(2) TaxBuilder supports multi-dimensional classification based on topic, style, domain, and programming language, whereas EvalTree organizes queries solely by model capability. 
(3) Our method fully leverages the power of advanced reasoning models, resulting in more accurate classifications. In contrast, EvalTree relies on sentence embeddings—which may lack precision—and K-Means clustering, with cluster labels generated by LLMs. 
(4) We provide analysis tools to identify model strengths and weaknesses, which are absent in EvalTree. 
(6) We also addresses sample size requirements for reliable evaluation within subcategories, an important factor not considered by EvalTree. 

\section{Experiments}
\label{sec:supplementary_material_of_experiment}

\subsection{Experimental Setup for Evaluation Method}
\label{sec:supplementary_material_of_experiment_evaluation_method}

\textbf{Models.} The PC$^{2}$ pointwise evaluation involves two types of model. In the pre-comparison stage, multiple models are utilized to generate responses. Here we employ gpt-4o-2024-11-20~\cite{openai2024gpt4o}, doubao-1-5-pro-32k-250115~\cite{doubao2024pro}, and Deepseek-V3-250324~\cite{deepseekai2024deepseekv3technicalreport} to ensure the diversity of the responses. For judge model, we adopt Deepseek-R1~\cite{deepseekai2025deepseekr1incentivizingreasoningcapability} due to its superior reasoning performance. 
Unless otherwise specified, we use the officially recommended decoding parameters. 

\noindent\textbf{Dataset.}
All our experiments are based on our pairwise human preference datasets SCAN-HPD, RewardBench-v2~\cite{malik2025rewardbench} and RMB~\cite{zhou2025rmb}. Pairwise human preference datasets usually contain a prompt and two responses, and the model's task is to determine which response is better. In pointwise evaluation, the judge model assigns scores to both responses, with the higher-scoring response being chosen as the winner. 

SCAN-HPD is a high-quality human preference dataset containing 636 samples. Queries are carefully filtered, and responses are annotated with majority human labels, making the dataset more representative of human preferences. A detailed description of SCAN-HPD is provided in \S~\ref{sec:appendix_evaluation_dataset_hpd}. 
RewardBench-v2 is a widely used benchmark for evaluating judge models, covering several domains: factuality, precise instruction following, math, safety, and focus. It is commonly used to measure and compare the performance of different language models. 
RMB is another recently popular large-scale human preference dataset comprising 10K queries for helpfulness and 7K for harmlessness. We only use the helpfulness subset in paper. 

Unless otherwise specified, our experiments are conducted using our own constructed dataset, SCAN-HPD, which will be publicly released alongside SCAN-D-V0. Results on RewardBench-v2 and RMB are presented only in \S~\ref{sec:appendix_pc2_exp}. 

\noindent\textbf{Baseline.} We employ naive pointwise evaluation as main baseline. 
Additionally, we also tested the impact of various components on P$^{2}$C pointwise evaluation. 
Direct metric decomposition refers to generating query-specific criteria based only on the query, without the process of pre-comparison~\cite{li2024fb, qin2024infobench, gupta2024unveiling}. 
Metric decomposition (single model) refers to using only gpt-4o with a temperature of 1, generating three responses. 
Metric decomposition (diverse model) involves using gpt-4o, doubao-1-5-pro, and Deepseek-V3 as models for generating auxiliary responses. Metric decomposition (diverse model) does not adopt the baseline guidance mechanism. 
For the baseline answer, we use gpt-4o-2024-11-20 with a temperature of 1.0 and top-p set to 0.95, generating a single response as the baseline answer. Regardless of whether we use naive pointwise evaluation or our method, when adding the baseline answer, we first evaluate it using the method without the baseline answer and include the evaluation result in the prompt. This process helps standardize the evaluation. Detailed instructions can be found in the respective prompts. 

\noindent\textbf{Prompt.} 
For PC$^2$ pointwise evaluation, the prompt for pre-comparison-derived criteria extraction is shown in Fig.~\ref{fig:prompt_pc2_criteria_generation}, the prompt for evaluation execution is shown in Fig.~\ref{fig:prompt_pc2_evaluation_execution}, the prompt for evaluation execution with baseline answer is shown in Fig.~\ref{fig:prompt_pc2_evaluation_execution_with_baseline}. 
For pairwise evaluation and naive pointwise evaluation, we directly use the prompts from MT-Bench~\cite{zheng2023judging}. 
For direct metric decomposition, the corresponding prompt is displayed in Fig.~\ref{fig:prompt_ncd}.

\begin{figure*}[!h]
\centering
\begin{promptbox}[PC$^2$ Pointwise Evaluation (Pre-Comparison-derived Criteria Extraction)]{blue_dist}
\small
You are an impartial judge responsible for evaluating the quality of responses provided by different LLMs to a given [question]. Your task is to design a comprehensive evaluation framework that includes clearly defined metrics and their respective weights. You shuold answer step by step. You should answer in English. Please carefully follow these steps: \\ 
 \\ 
1. **Analyze Responses**: You must first compare several provided [answers] and identify their differences. The objective of this comparison is to pinpoint distinguishing factors that significantly influence the quality of the responses. \\ 
2. **Develop Metrics**: Establish a hierarchical set of evaluation metrics. There should be 3 to 9 primary metrics. Each primary metric should have several detailed sub-metrics to provide specific, measurable criteria for evaluating the responses. \\ 
3. **Assign Weights**: Allocate appropriate weights to each metric based on its relative importance in distinguishing the quality of the responses. The weights should be integers, and the sum of all weights should equal 100. \\ 
4. **Output Format**: Present the final evaluation framework in a structured list format. You do not need to include the primary metrics; only the secondary metrics are required, in the following format: \\ 
<Evaluation\_Framework> \\ 
1. Description of Secondary Metric 1 | Weight 1 \\ 
2. Description of Secondary Metric 2 | Weight 2 \\ 
3. Description of Secondary Metric 3 | Weight 3 \\ 
... \\ 
<Evaluation\_Framework> \\ 
 \\ 
{[User Question]} \\ 
\{question\} \\ 
 \\ 
{[The Start of Assistant 1’s Answer]} \\ 
\{answer\_1\} \\ 
{[The End of Assistant 1’s Answer]} \\ 
 \\ 
{[The Start of Assistant 2’s Answer]} \\ 
\{answer\_2\} \\ 
{[The End of Assistant 2’s Answer]} \\ 
 \\ 
{[The Start of Assistant 3’s Answer]} \\ 
\{answer\_3\} \\ 
{[The End of Assistant 3’s Answer]}
\end{promptbox}
\caption{Prompt for pre-comparison-derived criteria extraction.}
\label{fig:prompt_pc2_criteria_generation}
\end{figure*}

\begin{figure*}[!h]
\centering
\begin{promptbox}[PC$^2$ Pointwise Evaluation (Evaluation Execution)]{blue_dist}
\small
Please act as an impartial judge and evaluate the quality of the response provided by an AI assistant to the user question displayed below. Your evaluation should consider the evaluation system displayed below. Begin your evaluation by providing an explanation for each metric, assessing the response objectively. Score each metric on a scale of 1 to 3, where 1 represents complete failure to meet the criterion and 3 represents perfection. The weights of the metrics must sum to 100, and the final weighted score should be calculated on a scale up to 300, reflecting the weighted sum of the individual scores. After evaluation, you must summarize the results within <The Start of Evaluation Result> and <The End of Evaluation Result>. Below is an example output: \\ 
<The Start of Evaluation Result> \\ 
Metric 1 | score: [2] \\ 
Metric 2 | score: [3] \\ 
Metric 3 | score: [1] \\ 
... \\ 
 \\ 
Final Weighted Score: [[200]] \\ 
<The End of Evaluation Result> \\ 
 \\ 
 \\ 
{[Question]} \\ 
\{question\} \\ 
 \\ 
{[The Start of Evaluation System]} \\ 
\{eval\_system\} \\ 
{[The End of Evaluation System]} \\ 
 \\ 
{[The Start of Assistant’s Answer]} \\ 
\{answer\} \\ 
{[The End of Assistant’s Answer]}
\end{promptbox}
\caption{Prompt for evaluation execution.}
\label{fig:prompt_pc2_evaluation_execution}
\end{figure*}

\begin{figure*}[!h]
\centering
\begin{promptbox}[PC$^2$ Pointwise Evaluation (Evaluation Execution with Baseline Answer)]{blue_dist}
\small
Please act as an impartial judge and evaluate the quality of the response provided by an AI assistant to the user question displayed below. Your evaluation should consider the evaluation system displayed below. Begin your evaluation by providing an explanation for each metric, assessing the response objectively. Score each metric on a scale of 1 to 3. Use the baseline answer as a baseline; a higher score indicates a better response compared to the baseline answer, while a lower score indicates a worse response. The weights of the metrics must sum to 100, and the final weighted score should be calculated on a scale of 100 to 300, reflecting the weighted sum of the individual scores. After evaluation, you must summarize the results within <The Start of Evaluation Result> and <The End of Evaluation Result>. Below is an example output: \\ 
<The Start of Evaluation Result> \\ 
Metric 1 | score: [2] \\ 
Metric 2 | score: [3] \\ 
Metric 3 | score: [1] \\ 
... \\ 
 \\ 
Final Weighted Score: [[200]] \\ 
<The End of Evaluation Result> \\ 
 \\ 
 \\ 
{[Question]} \\ 
\{question\} \\ 
 \\ 
{[The Start of Evaluation System]} \\ 
\{eval\_system\} \\ 
{[The End of Evaluation System]} \\ 
 \\ 
{[The Start of Baseline Answer]} \\ 
\{answer\_baseline\} \\ 
{[The End of Baseline Answer]} \\ 
 \\ 
{[The Start of Evaluation for Baseline Answer]} \\ 
\{critic\_baseline\} \\ 
{[The End of Evaluation for Baseline Answer]} \\ 
 \\ 
{[The Start of Assistant’s Answer]} \\ 
\{answer\} \\ 
{[The End of Assistant’s Answer]}
\end{promptbox}
\caption{Prompt for evaluation execution with baseline answer.}
\label{fig:prompt_pc2_evaluation_execution_with_baseline}
\end{figure*}

\begin{figure*}[!h]
\centering
\begin{promptbox}[Prompt of Naive Criteria Decomposition Pointwise Evaluation]{medium_gray}
\small
You are an impartial judge responsible for evaluating the quality of responses provided by different LLMs to a given [question]. Your task is to design a comprehensive evaluation framework that includes clearly defined metrics and their respective weights. You shuold answer step by step. You should answer in English. Please carefully follow these steps: \\ 
 \\ 
1. **Analyze Question**: You must first analyze the question. The objective of this comparison is to find important factors that significantly influence the quality of the responses. \\ 
2. **Develop Metrics**: Establish a hierarchical set of evaluation metrics. There should be 3 to 9 primary metrics. Each primary metric should have several detailed sub-metrics to provide specific, measurable criteria for evaluating the responses. \\ 
3. **Assign Weights**: Allocate appropriate weights to each metric based on its relative importance in distinguishing the quality of the responses. The weights should be integers, and the sum of all weights should equal 100. \\ 
4. **Output Format**: Present the final evaluation framework in a structured list format. You do not need to include the primary metrics; only the secondary metrics are required, in the following format: \\ 
<Evaluation\_Framework> \\ 
1. Description of Secondary Metric 1 | Weight 1 \\ 
2. Description of Secondary Metric 2 | Weight 2 \\ 
3. Description of Secondary Metric 3 | Weight 3 \\ 
... \\ 
<Evaluation\_Framework> \\ 
 \\ 
{[User Question]} \\ 
\{question\}
\end{promptbox}
\caption{Prompt for naive criteria decomposition.}
\label{fig:prompt_ncd}
\end{figure*}

\subsection{More Experiment on PC\texorpdfstring{$^{2}$}{2}-Pointwise Evaluation}
\label{sec:appendix_pc2_exp}

\noindent\textbf{More Benchmarks.} To further validate the effectiveness of our method, we conduct experiments on additional benchmarks, namely RewardBench-v2~\cite{malik2025rewardbench} and RMB~\cite{zhou2025rmb}. Both benchmarks adopt the Best-of-N evaluation paradigm, which requires judge model selecting the best answer among multiple candidates, making the task inherently more difficult. 
The experimental results are reported in Tab.~\ref{tab:more_benchmarks}. As can be seen, our method consistently improves performance across all benchmarks, thereby validating its broad effectiveness. 

\begin{table}[!h]
    \centering
    \begin{tabular}{lcc}
        \toprule
        \textbf{} & \textbf{naive} & \textbf{ours} \\
        \midrule
        RMB            & 0.4068 & 0.5969 \\
        RewardBench-v2 & 0.5551 & 0.5869 \\
        \bottomrule
    \end{tabular}
    \caption{Evaluation accuracy results on RewardBench-v2 and RMB.}
    \label{tab:more_benchmarks}
\end{table}

\noindent\textbf{Ablation Study.} To gain deeper insights into the factors that contribute to the performance of our method, we further conduct extensive ablation studies along three dimensions: (\emph{i}) pre-com model combination, (\emph{ii}) number of pre-com models, and (\emph{iii}) order of pre-com models. The results are summarized in Tab.~\ref{tab:ablation_combination}, \ref{tab:ablation_number}, and \ref{tab:ablation_order}, respectively. 

\noindent\textbf{(1) Ablation on model combination.} For model combination (Tab.~\ref{tab:ablation_combination}), we observe that multi-model setups consistently outperform the single-model baseline (\texttt{gpt-4o}), confirming that a single model is insufficient for robust judging performance. However, the differences among various multi-model combinations are relatively small, indicating that the \emph{diversity} of the incorporated models, rather than the different combinations of models, is the key factor that drives performance improvements. 

\noindent\textbf{(2) Ablation on number of models.} As shown in Tab.~\ref{tab:ablation_number}, increasing the number of models does not always lead to better performance. The best result is obtained when using $3$ models, whereas adding more models sometimes introduces redundancy or noise, resulting in a slight drop in accuracy. 

\noindent\textbf{(3) Ablation on order of models.} Finally, we investigate the effect of model ordering (Tab.~\ref{tab:ablation_order}). The results show that different ordering permutations lead to only marginal differences in accuracy, suggesting that the impact of model order is relatively limited. This indicates that our approach is robust to changes in execution sequence, and the overall performance is primarily determined by the chosen set of models rather than their specific order. 

\noindent These findings provide practical guidance for constructing and optimizing multi-model judging systems, highlighting the importance of not only choosing the right set of models but also carefully designing their arrangement.

\begin{table*}[!h]
\centering
\begin{tabular}{lc}
    \toprule
    \textbf{Model Combination} & \textbf{ACC} \\
    \midrule
    single model (\texttt{gpt-4o}) & 0.5974 \\
    gpt-4o + doubao-pro-1.5-32k + Deepseek-V3 & 0.6962 \\
    gpt-4o + doubao-pro-1.5-32k + Qwen2.5-7B-Instruct & 0.6904 \\
    gpt-4o + doubao-pro-1.5-32k + Meta-Llama-3.1-8B-Instruct & 0.6747 \\
    gpt-4o + doubao-pro-1.5-32k + Phi-4-mini-instruct & \textbf{0.7077} \\
    gpt-4o + Qwen2.5-7B-Instruct + Meta-Llama-3.1-8B-Instruct & 0.6825 \\
    gpt-4o + Qwen2.5-7B-Instruct + Phi-4-mini-instruct & 0.6574 \\
    Qwen2.5-7B-Instruct + Phi-4-mini-instruct + Meta-Llama-3.1-8B-Instruct & 0.6699 \\
    \bottomrule
\end{tabular}
\caption{Ablation on model combination.}
\label{tab:ablation_combination}
\end{table*}

\begin{table*}[!h]
\centering
\resizebox{0.8\textwidth}{!}{
    \begin{tabular}{clc}
    \toprule
    \textbf{Number of Models} & \textbf{Model Combination} & \textbf{ACC} \\
    \midrule
    2 & gpt-4o + doubao-pro-1.5-32k & 0.6699 \\
    3 & gpt-4o + Deepseek-V3 + doubao-pro-1.5-32k & \textbf{0.6962} \\
    4 & gpt-4o + doubao-pro-1.5-32k + Deepseek-V3 + Qwen2.5-7B-Instruct & 0.6621 \\
    5 & gpt-4o + doubao-pro-1.5-32k + Deepseek-V3 + Qwen2.5-7B-Instruct + Phi-4-mini-instruct & 0.6448 \\
    \bottomrule
    \end{tabular}}
\caption{Ablation on number of models.}
\label{tab:ablation_number}
\end{table*}

\begin{table*}[!h]
\centering
\begin{tabular}{lc}
    \toprule
    \textbf{Order of Models} & \textbf{ACC} \\
    \midrule
    gpt-4o → doubao-pro-1.5-32k → Deepseek-V3 & 0.6668 \\
    gpt-4o → Deepseek-V3 → doubao-pro-1.5-32k & \textbf{0.6962} \\
    Deepseek-V3 → gpt-4o → doubao-pro-1.5-32k & 0.6699 \\
    Deepseek-V3 → doubao-pro-1.5-32k → gpt-4o & 0.6920 \\
    doubao-pro-1.5-32k → gpt-4o → Deepseek-V3 & 0.6778 \\
    doubao-pro-1.5-32k → Deepseek-V3 → gpt-4o & 0.6605 \\
    \bottomrule
\end{tabular}
\caption{Ablation on order of models.}
\label{tab:ablation_order}
\end{table*}

\subsection{Experimental Setup for Evaluation and Leaderboard}
\label{sec:appendix_exp_setup_leaderboard}

\textbf{Models.} We select 21 mainstream LLMs for evaluation. 
For closed-source models, we select claude3.7-sonnet-20250219~\cite{anthropic2025claude37}, gpt-4o-2024-11-20~\cite{openai2024gpt4o}, qwen-max-2024-10-15~\cite{}, doubao-1-5-pro-32k-250115~\cite{doubao2024pro}, and hunyuan-standard-2025-02-10~\cite{hunyuan2025}. 
For open-source models, we select Deepseek-V3-250324~\cite{deepseekai2024deepseekv3technicalreport}, Deepseek-R1-250120~\cite{deepseekai2025deepseekr1incentivizingreasoningcapability}, QwQ-32B~\cite{qwq32b}, Gemma-3-27B-IT~\cite{team2025gemma}, Qwen2.5-72B-Instruct~\cite{yang2024qwen2}, Gemma-3-4B-IT~\cite{team2025gemma}, Qwen2.5-32B-Instruct~\cite{yang2024qwen2}, Meta-Llama-3.1-70B-Instruct~\cite{grattafiori2024llama}, Qwen2.5-7B-Instruct~\cite{yang2024qwen2}, Meta-Llama-3.1-8B-Instruct~\cite{grattafiori2024llama}, Mistral-7B-Instruct-v0.3~\cite{}, and Phi-4-Mini-Instruct~\cite{abdin2024phi}. 
Additionally, we evaluate recent open-source models: GPT-OSS-20B~\cite{openai2025gptoss120bgptoss20bmodel}, GPT-OSS-120B~\cite{openai2025gptoss120bgptoss20bmodel}, Qwen3-8B~\cite{yang2025qwen3technicalreport}, and Qwen3-32B~\cite{yang2025qwen3technicalreport}. 
Among these, Deepseek-V3-250324 and Deepseek-R1-250120 are deployed with fp8, while the other models are deployed with bf16. 
For all models, if their official decoding parameters are provided, we use the default settings. If not, we set the temperature to 0.7 and top\_p to 0.95. 

\subsection{Leaderboard}
\label{sec:leaderboard}

We propose in the paper how to quickly check the weaknesses of a LLM through the tree structure. Due to the limitation of the paper's length, we only present the first-level nodes (i.e., the six domain) in the paper, while the more fine-grained results are presented on our project website. The leaderboard results, as presented in Tab.~\ref{tab:leaderboard}, reveal some valuable weaknesses of existing LLMs. These weaknesses can significantly guide the model's development if they are timely available during the development process of the model.

\noindent The new overall leader, GPT-OSS-120B, exhibits a strong aptitude for technical tasks, securing top ranks in coding, reasoning, and mathematics. However, its performance is less dominant in creative domains like roleplay. A clear trade-off is visible in other models; for example, Deepseek-R1-250120 excels more in writing and roleplay but is weaker in mathematics. Conversely, Gemma-3-27B-IT achieves the top score in knowledge and ranks highly in reasoning, yet its performance is notably lower in STEM-oriented subjects like coding and mathematics. Despite a lower overall ranking, claude-3-7-sonnet-20250219 shows competitive coding abilities, but its weaker reasoning skills hinder its general performance. Additionally, it is worth mentioning that doubao-1-5-pro-32k-250115 possesses a notable advantage in mathematical abilities relative to its overall rank.

\begin{table*}[!t]
\centering
\resizebox{1.0\textwidth}{!}{
\begin{tabular}{lccccccc}
\toprule
\textbf{Model} & \textbf{Overall} & \textbf{Writing} & \textbf{Roleplay} & \textbf{Knowledge} & \textbf{Coding} & \textbf{Mathematics} & \textbf{Reasoning} \\
\midrule
\multicolumn{8}{c}{\textbf{Closed-source and Open-source (>100B) LLMs}} \\
\midrule
GPT-OSS-120B & \cellcolor{green_first}242.24(1) & 233.59(4) & 231.33(7) & \cellcolor{green_third}247.99(3) & \cellcolor{green_first}263.35(1) & \cellcolor{green_second}241.59(2) & \cellcolor{green_first}235.70(1) \\
Deepseek-V3-250324 & \cellcolor{green_third}237.65(3) & 233.04(5) & 237.41(5) & \cellcolor{green_second}250.91(2) & \cellcolor{green_third}247.09(3) & 220.01(5) & 229.93(4) \\
Deepseek-R1-250120 & 235.25(4) & \cellcolor{green_first}244.14(1) & \cellcolor{green_second}248.29(2) & 238.19(4) & 231.94(6) & 190.58(8) & \cellcolor{green_second}234.91(2) \\
claude-3-7-sonnet-20250219 \faLock{} & 225.07(7) & 232.74(6) & 224.94(10) & 229.79(8) & 238.37(5) & 187.36(10) & 203.56(11) \\
gpt-4o-2024-11-20 \faLock{} & 214.80(9) & 225.90(8) & 234.42(6) & 219.31(10) & 202.03(10) & 171.82(11) & 210.50(9) \\
qwen-max-2024-10-15 \faLock{} & 182.59(13) & 187.44(13) & 174.14(14) & 192.56(13) & 183.33(13) & 164.28(13) & 178.68(14) \\
doubao-1-5-pro-32k-250115 \faLock{} & 182.23(14) & 175.36(18) & 173.19(15) & 192.40(14) & 187.05(12) & 192.35(6) & 181.46(13) \\
hunyuan-standard-2025-02-10 \faLock{} & 156.44(19) & 163.21(20) & 144.43(20) & 156.85(18) & 158.49(18) & 149.47(15) & 152.47(19) \\
\midrule
\multicolumn{8}{c}{\textbf{Open-source (<100B) LLMs}} \\
\midrule
Qwen3-32B & \cellcolor{green_second}241.13(2) & \cellcolor{green_second}241.29(2) & \cellcolor{green_third}246.03(3) & 234.47(6) & 246.87(4) & \cellcolor{green_first}245.29(1) & 224.93(6) \\
Gemma-3-27B-IT & 229.88(5) & \cellcolor{green_third}236.63(3) & 243.59(4) & \cellcolor{green_first}253.25(1) & 207.01(9) & 192.19(7) & \cellcolor{green_third}233.82(3) \\
QwQ-32B & 229.38(6) & 232.26(7) & \cellcolor{green_first}250.41(1) & 237.71(5) & 224.73(7) & 187.67(9) & 228.24(5) \\
Qwen3-8B & 222.88(8) & 224.63(9) & 227.69(9) & 210.07(11) & 219.59(8) & \cellcolor{green_third}241.40(3) & 216.47(7) \\
GPT-OSS-20B & 213.72(10) & 196.23(12) & 176.09(13) & 230.79(7) & \cellcolor{green_second}251.59(2) & 225.26(4) & 212.92(8) \\
Gemma-3-4B-IT & 203.43(11) & 224.30(10) & 230.57(8) & 219.62(9) & 167.27(15) & 137.90(17) & 207.15(10) \\
Qwen2.5-72B-Instruct & 190.42(12) & 197.99(11) & 179.91(11) & 194.18(12) & 190.53(11) & 171.61(12) & 194.16(12) \\
Qwen2.5-32B-Instruct & 176.10(15) & 185.19(14) & 170.58(16) & 174.20(15) & 174.39(14) & 162.31(14) & 173.53(15) \\
Meta-Llama-3.1-70B-Instruct & 170.51(16) & 183.13(15) & 178.49(12) & 167.73(16) & 165.25(16) & 133.37(18) & 170.07(16) \\
Qwen2.5-7B-Instruct & 166.67(17) & 179.10(16) & 164.64(17) & 162.77(17) & 161.99(17) & 146.67(16) & 163.15(17) \\
Meta-Llama-3.1-8B-Instruct & 159.01(18) & 176.65(17) & 159.26(18) & 152.30(20) & 154.93(19) & 119.67(20) & 159.35(18) \\
Mistral-7B-Instruct-v0.3 & 148.94(20) & 163.44(19) & 158.07(19) & 152.55(19) & 138.24(21) & 114.74(21) & 133.61(21) \\
Phi-4-mini-instruct & 143.99(21) & 150.48(21) & 144.18(21) & 142.36(21) & 143.21(20) & 123.93(19) & 148.08(20) \\
\bottomrule
\end{tabular}}
\caption{Overall and domain-level scores (fine-grained results on project page). }
\label{tab:leaderboard}
\end{table*}

\section{Taxonomy and Dataset}
\label{sec:supplementary_material_of_dataset}

\subsection{TaxBuilder}
\label{sec:appendix_TaxBuilder}

\subsubsection{Manually Initialized Basic Taxonomy Tree}
\label{sec:appendix_TaxBuilder_basic_tree}

TaxBuilder requires a manually crafted initial taxonomy as a starting point. This step is quite straightforward. First, we need to define a root node—typically a domain—such as coding. 

Next, to capture the various aspects of this domain, TaxBuilder asks users to manually define several classification principles, such as task types, technical domains, and programming languages. 

\begin{figure*}[!h]
\centering
    \includegraphics[width=0.9\textwidth]{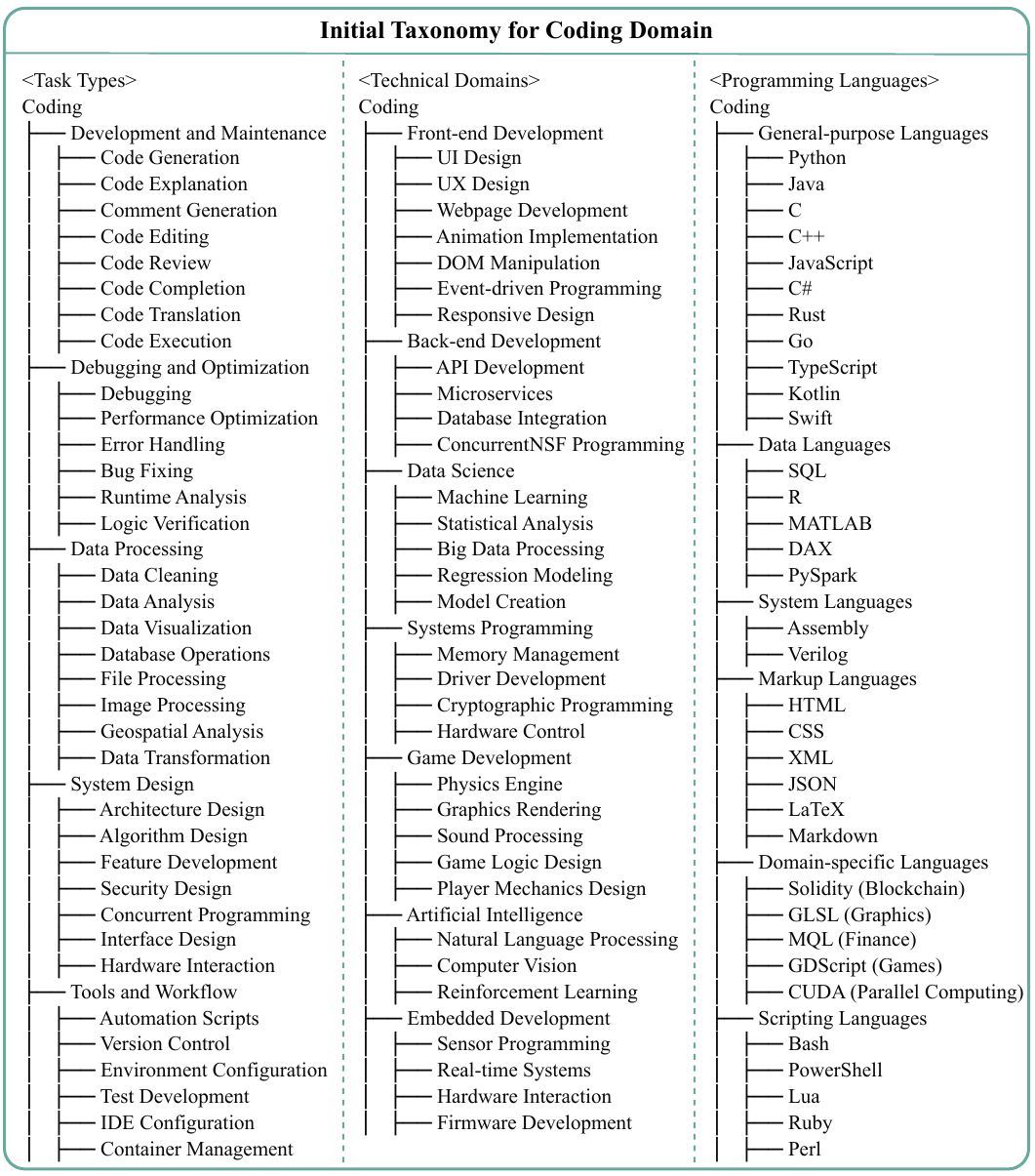}
\caption{Initial taxonomy for coding domain.}
\label{fig:init_tree_coding}
\end{figure*}

Finally, we manually construct a basic fine-grained taxonomy. Fig.~\ref{fig:init_tree_coding} shows an example of a manually created taxonomy for the coding domain. Additional examples can be found in our open-source code repository. 

\subsubsection{Initial Annotation for Query Type Tags}
\label{sec:appendix_TaxBuilder_init_tags}

We use gpt-4o-2024-11-20 to generate a large number of low-quality query type tags. The prompt used for generation is shown in Fig.~\ref{fig:initial_annotation_for_tags}. 

\begin{figure*}[!h]
\centering
\begin{promptbox}[Initial low-quality Type Labeling]{medium_gray}
\tiny
\#\#\# Task \\
Your task is to categorize a given query based on the following rules. For each rule, explain your reasoning before assigning a label. Evaluate the query against each domain in the order provided below until a match is found. Once a domain is identified, proceed with its specific sub-rules. If no domain matches, proceed to the next domain check. If no specific domain applies, return a generic \texttt{"other"} label.\\
\#\#\#\# Rules \\
\#\#\#\#\# \*\*Step 1: Determine the Domain\*\* \\
Evaluate the query against the following domains in this order. Stop and proceed with the corresponding sub-rules once a domain is matched. If the domain is labeled \texttt{"other"}, return \texttt{\{"domain": "other", "domain\_name": "\textless custom\_domain\_name\textgreater "\}} and do not proceed further. Explain your reasoning before assigning a label.\\
1. \*\*Writing\*\* 
   - Condition: If the query relates to writing, purposeful professional writing, literature, storytelling, grammar, or language-related topics. 
   - Label: \texttt{"writing"} 
   - If not matched, proceed to the next domain.

2. \*\*Roleplay\*\* 
   - Condition: If the query involves roleplay, character interactions, storytelling, or immersive scenario-based dialogue.
   - Label: \texttt{"roleplay"}
   - If not matched, proceed to the next domain.

3. \*\*Coding\*\* 
   - Condition: If the query relates to programming, algorithms, or code-related topics.
   - Label: \texttt{"coding"}
   - If not matched, proceed to the next domain.

4. \*\*Mathematics\*\* 
   - Condition: If the query pertains to mathematical concepts, computations, proofs, or problem-solving.
   - Label: \texttt{"mathematics"}
   - If not matched, proceed to the next domain.

5. \*\*Reasoning\*\* 
   - Condition: If the query pertains to reasoning, logic, critical thinking, or problem-solving without direct reference to specific knowledge or programming.
   - Label: \texttt{"reasoning"}
   - If not matched, proceed to the next domain.

6. \*\*Knowledge\*\* 
   - Condition: If the query pertains to factual knowledge or general subject areas such as science, history, literature, philosophy, or current affairs.
   - Label: \texttt{"knowledge"}
   - If not matched, label as \texttt{"other"} with a custom \texttt{"domain\_name"} and stop.

\#\#\#\# Step 2: Domain-Specific Categorization \\
Once the domain is determined, apply the corresponding sub-rules below. If the domain is \texttt{"other"}, skip this step. Explain your reasoning before assigning a label. \\
- \*\*Domain: "writing"\*\* 
  - \*\*Label 2: Writing Topic\*\* 
    - Specify the writing topic in lowercase (e.g., \texttt{"creative writing"}, \texttt{"speech"}, \texttt{"LLM prompt"}, \texttt{"game"}, \texttt{"academic"}, \texttt{"social media"}, \texttt{"shopping"}, \texttt{"formal"}, \texttt{"casual"}, etc.).
    - If unclear or not explicitly mentioned, use \texttt{"null"}.
    - If none of the predefined topics fit, provide a custom description.
  - \*\*Label 3: Task Type\*\* 
    - Choose applicable task types (multiple may apply):
      - \texttt{"Creative Writing"}: Crafting original content, poems, or creative pieces.
      - \texttt{"Polishing"}: Improving clarity, style, and flow.
      - \texttt{"Grammar Check"}: Identifying and correcting grammar mistakes.
      - \texttt{"Content Refining"}: Enhancing word choice, tone, or style.
      - \texttt{"Summarization"}: Condensing text while retaining the core message.
      - \texttt{"Expanding"}: Adding details or elaboration.
      - \texttt{"Paraphrasing"}: Restating content differently.
      - \texttt{"Writing Feedback"}: Providing constructive feedback.
      - \texttt{"Translating"}: Converting text between languages.
      - \texttt{"Tone Adjustment"}: Modifying tone for audience/context.
      - \texttt{"Text Summarization"}: Analyzing meaning, themes, or key points.
    - If none fit, provide a custom description.

- \*\*Domain: "roleplay"\*\* 
  - \*\*Label 2: Genre\*\* 
    - Specify the genre in lowercase (e.g., \texttt{"sci-fi"}, \texttt{"fantasy"}, \texttt{"historical"}, \texttt{"romance"}, \texttt{"mystery"}, \texttt{"horror"}, \texttt{"action/adventure"}, \texttt{"supernatural"}, etc.).
    - If multiple genres apply, list them all.
    - If none fit, provide a custom description.
  - \*\*Label 3: Task Type\*\* 
    - Choose applicable task types (multiple may apply):
      - \texttt{"Dialogue Generation"}: Creating natural character conversations.
      - \texttt{"Content Creation"}: Writing from a character’s perspective.
      - \texttt{"Question Answering"}: Responding within character constraints.
      - \texttt{"Emotional Simulation"}: Modeling character-appropriate emotions.
      - \texttt{"Decision Support"}: Providing character-based guidance.
      - \texttt{"Skill Training"}: Simulating expert/practice partners.
      - \texttt{"Behavior Simulation"}: Replicating character behaviors.
      - \texttt{"Evaluation \& Feedback"}: Assessing from a character’s perspective.
    - If none fit, provide a custom description.

- \*\*Domain: "coding"\*\* 
  - \*\*Label 2: Programming Language\*\* 
    - Specify the language in lowercase (e.g., \texttt{"python"}, \texttt{"java"}, \texttt{"c++"}, \texttt{"sql"}, etc.).
    - If unclear or not mentioned, use \texttt{"null"}.
  - \*\*Label 3: Task Type\*\* 
    - Choose applicable task types (multiple may apply):
      - \texttt{"Code Generation"}: Creating code from a prompt.
      - \texttt{"Debugging"}: Resolving code errors.
      - \texttt{"Code Explanation"}: Explaining code functionality.
      - \texttt{"Code Editing"}: Modifying existing code.
      - \texttt{"Conceptual QA"}: Answering programming theory questions.
      - \texttt{"Code Translation"}: Converting code between languages.
      - \texttt{"Bug Fixing"}: Fixing specific code issues.
      - \texttt{"Testing"}: Writing/analyzing tests.
      - \texttt{"Data Structures \& Algorithms"}: Solving related problems.
      - \texttt{"System Design"}: Designing software architecture.
      - \texttt{"Optimization"}: Improving code efficiency.
      - \texttt{"Security"}: Addressing vulnerabilities.
      - \texttt{"Documentation"}: Writing/improving documentation.
    - If none fit, provide a custom description.

- \*\*Domain: "mathematics"\*\* 
  - \*\*Label 2: Subfield\*\* 
    - Specify the subfield in lowercase (e.g., \texttt{"algebra"}, \texttt{"calculus"}, \texttt{"geometry"}, \texttt{"statistics \& probability"}, etc.).
    - If unclear, use \texttt{"null"}.
  - \*\*Label 3: Task Type\*\* 
    - Choose applicable task types (multiple may apply):
      - \texttt{"Computation"}: Performing calculations.
      - \texttt{"Proof Writing"}: Writing formal proofs.
      - \texttt{"Conceptual Understanding"}: Explaining concepts.
      - \texttt{"Problem-Solving"}: Applying techniques to challenges.
      - \texttt{"Real-World Application"}: Connecting math to scenarios.
      - \texttt{"Graphing \& Visualization"}: Interpreting/drawing graphs.
      - \texttt{"Logical Deduction"}: Deriving conclusions logically.
    - If none fit, provide a custom description.

- \*\*Domain: "reasoning"\*\* 
  - \*\*Label 2: Reasoning Type\*\* 
    - Specify the type in lowercase (e.g., \texttt{"logical reasoning"}, \texttt{"critical thinking"}, \texttt{"problem-solving"}, \texttt{"analytical thinking"}, etc.).
    - If unclear, use \texttt{"null"}.
  - \*\*Label 3: Task Type\*\* 
    - Choose applicable task types (multiple may apply):
      - \texttt{"Logical Deduction"}: Deriving conclusions logically.
      - \texttt{"Puzzle Solving"}: Solving riddles or brainteasers.
      - \texttt{"Argument Evaluation"}: Analyzing argument validity.
      - \texttt{"Scenario Analysis"}: Considering scenario outcomes.
      - \texttt{"Theoretical Reasoning"}: Exploring abstract concepts.
      - \texttt{"Creative Problem-Solving"}: Finding novel solutions.
      - \texttt{"Ethical Reasoning"}: Evaluating based on ethics.
    - If none fit, provide a custom description.

- \*\*Domain: "knowledge"\*\* 
  - \*\*Label 2: Subject\*\* 
    - Specify the subject in lowercase (e.g., \texttt{"history"}, \texttt{"philosophy"}, \texttt{"geography"}, \texttt{"medicine"}, \texttt{"economics"}, etc.).
    - If unclear, use \texttt{"null"}.
  - \*\*Label 3: Task Type\*\* 
    - Choose applicable task types (multiple may apply):
      - \texttt{"Fact Recall"}: Retrieving specific facts.
      - \texttt{"Conceptual Understanding"}: Explaining/interpreting ideas.
      - \texttt{"Comparative Analysis"}: Comparing concepts.
      - \texttt{"Causal Reasoning"}: Examining cause-and-effect.
      - \texttt{"Logical Deduction"}: Applying logic to conclude.
      - \texttt{"Hypothetical Thinking"}: Exploring "what if" scenarios.
      - \texttt{"Problem-Solving"}: Solving practical/theoretical problems.
      - \texttt{"Opinion-Based"}: Providing subjective reasoning.
    - If none fit, provide a custom description.

\#\#\#\# Step 3: Final Output \\
Return a Python dictionary based on the identified domain and labels:\\
- For \texttt{"other"}: \texttt{\{"domain": "other", "domain\_name": "\textless custom\_domain\_name\textgreater "\}}\\
- For specific domains: \\
\texttt{\{"domain": "\textless domain\textgreater ", "task\_type": ["\textless task1\textgreater ", "\textless task2\textgreater "], "content\_type": ["\textless content1\textgreater ", "\textless content2\textgreater "]\}}\\
\quad - Replace \texttt{\textless domain\textgreater } with the identified domain (e.g., \texttt{"writing"}).\\
\quad - Replace \texttt{"task"} with the list of identified task types (or empty list \texttt{[]} if none apply).\\
\quad - Replace \texttt{\textless content\textgreater } with the corresponding value (e.g., \texttt{"poetry"}, \texttt{"sci-fi"}, \texttt{"history"}, etc.).\\
\end{promptbox}
\caption{Prompt for initial annotation of tags.}
\label{fig:initial_annotation_for_tags}
\end{figure*}

\newpage
\subsubsection{Algorithm for TaxBuilder}
\label{sec:appendix_TaxBuilder_alg}

The complete TaxBuilder algorithm is presented in Alg.~\ref{alg:tax_builder}. 

\begin{algorithm}[H]
\caption{TaxBuilder Algorithm}
\label{alg:tax_builder}
\begin{algorithmic}[1]
\State \textbf{Input:} Initial tree $T_{init}$, new nodes $TG_{init}$, LLM-as-Decision-Maker of node insertion \texttt{DM$_{ins}$}
\State \textbf{Output:} Final taxonomy tree $T_{final}$
\\
\Function{InsertNode}{$T, tg$}
    \State $d \gets \texttt{DM$_{ins}$}(tg, T^{c})$
    \If{$d = $ "<E>"}
        \State \Return $T$
    \ElsIf{$d = $ "<S>"}
        \State $T^{c} \gets T^{c} \cup \{tg\} $
        \State \Return $T$
    \Else
        \Comment{Child level: LLM returns parent node name}
        \State $T^{c_{i}} \gets d$
        \State \Return \Call{InsertNode}{$T^{c_{i}}, tg$}
    \EndIf
\EndFunction
\\
\Function{TaxBuilder}{$T_{init}$, $TG_{init}$}
    \State $T_{cur} \gets T_{init}$
    \ForAll{$tg \in TG_{init}$}
        \State $T_{cur} \gets$ \Call{InsertNode}{$T_{cur}, tg$}
    \EndFor
    \State $T_{cur} \gets$ \Call{RefinePruneNode}{$T_{cur}$}
    \Comment{Node refinement and pruning}
    \State $T_{final} \gets$ \Call{PruneLayer}{$T_{cur}$}
    \Comment{Layer pruing}
    \State \Return $T_{final}$
\EndFunction

\end{algorithmic}
\end{algorithm}

\newpage
\subsubsection{Prompts for TaxBuilder}
\label{sec:appendix_TaxBuilder_prompt}

The prompt for node insertion in the LLM-as-Decision-Maker framework is shown in Fig.~\ref{fig:dm_ins}. The prompt for node refinement and pruning in the LLM-as-Decision-Maker framework is shown in Fig.~\ref{fig:dm_node_refine_prun}. 

\begin{figure*}[!h]
\centering
\begin{promptbox}[Prompt of LLM-as-Decision-Maker (Node Insertion)]{orange_dist}
\small
Your task is to add a new node to a tree-structured classification system. I will give you some \texttt{[Current Level Nodes]} and a \texttt{[New Node]}. You need to determine the next action and respond strictly in the specified format. Before giving the final answer, you must perform brief analysis. \\

\# Task 1 \\
Analyze whether the \texttt{[New Node]} shares identical meaning with any \texttt{[Current Level Nodes]} (including exact matches, case variations, abbreviations, or conceptually equivalent expressions). If identical, return \texttt{<decision>EXIST</decision>} and skip subsequent tasks. \\
Analyze whether the \texttt{[New Node]} represents a specific category or a classification criterion. If it's a classification criterion, return \texttt{<decision>EXIST</decision>} and skip subsequent tasks. \\

\# Task 2 \\
Analyze whether the \texttt{[New Node]} doesn't belong to any subordinate node of \texttt{[Current Level Nodes]} but should be at the same level. If yes, return \texttt{<decision>ADD</decision>} and skip subsequent tasks. \\

\# Task 3 \\
Identify which \texttt{[Current Level Node]} the \texttt{[New Node]} should belong to as a subordinate node. Return the parent node's name wrapped in \texttt{<decision></decision>}. \\

\#\# Example \\
Current Level Nodes: \texttt{["Fruits", "Vegetables", "Electronics"]} \\
New Node: Phone \\
Output: \texttt{<decision>Electronics</decision>} \\

\#\# New Node \\
\{new\_node\} \\

\#\# Current Level Nodes \\
\{current\_keys\} \\

\#\# Reference Classification System (for reference only) \\
\{init\_tree\}
\end{promptbox}
\caption{Prompt of LLM-as-Decision-Maker (node insertion).}
\label{fig:dm_ins}
\end{figure*}

\newpage

\begin{figure*}[!h]
\centering
\begin{promptbox}[Prompt of LLM-as-Decision-Maker (Node Refinement and Pruning)]{orange_dist}
\scriptsize
Your task is to optimize a given tree structure classification system. I will provide all the next-level nodes under a certain category (referred to as \texttt{"current layer nodes"}) and the leaf nodes among them (referred to as \texttt{"leaf nodes in the current layer"}). You need to determine the next steps and return results in the specified format. Before giving the final answer, please provide a brief analysis. Note: Do not modify the names of the original categories, and do not omit any categories from the \texttt{"current layer nodes"} when returning results. Each task analysis should analyze all categories without omission. \\

If you see that the \texttt{"current layer nodes"} are all classification criteria rather than specific categories, this layer absolutely must not be modified. \\

\#\#\# Task 1: Split \\
Examine all nodes in the \texttt{"leaf nodes in the current layer"} one by one, determine if there are any nodes that are not \texttt{"single concepts"} and split them. If the newly split category already exists, do not return that new category. A \texttt{"single concept"} refers to an independent concept, as opposed to a \texttt{"combined concept"} which is formed by combining multiple concepts. For example, \texttt{"mythology novel"} is a combined concept, consisting of \texttt{"mythology"} and \texttt{"novel"}. Return format: \\
\texttt{[(old\_key1, [new\_key1, new\_key2]), (old\_key2, [new\_key1, new\_key2])]} \\

\#\#\# Task 2: Parent-Child Relationship \\
Analyze all nodes in the \texttt{"current layer nodes"} one by one, determine if there exist parent-child relationships, or if multiple nodes can be merged into a parent node, and return the results. If there are no qualifying situations, do not return anything. Return format: \\
\texttt{[(parent\_key1, [old\_child\_key1, old\_child\_key2]), (parent\_key2, [old\_child\_key1, old\_child\_key2])]} \\

\#\#\# Task 3: Delete \\
Examine all nodes in the \texttt{"current layer nodes"} one by one, identify meaningless categories (such as null values, blanks, etc.) or extremely rare categories in the real world, and delete them. Return format: \\
\texttt{[old\_key\_to\_delete1, old\_key\_to\_delete2]} \\

\#\#\# Task 4: Merge \\
Analyze all nodes in the \texttt{"current layer nodes"} one by one, identify categories with consistent concepts and merge them. Return format: \\
\texttt{[(new\_key1, [old\_key\_to\_merge1, old\_key\_to\_merge2]), (new\_key2, [old\_key\_to\_merge1, old\_key\_to\_merge2])]} \\

\#\#\# Task 5: Keep \\
Preserve the remaining nodes. Return format: \\
\texttt{[old\_key1, old\_key2]} \\

\#\# Output Format \\
Provide a brief analysis for each task and give the results after the task is executed. 

After completing all tasks, format the final result as a Python dictionary wrapped in \texttt{<decision></decision>} tags, in the following format, where all keys with \texttt{old\_} must have appeared in the \texttt{"current layer nodes"}: \\
\texttt{<decision>} \texttt{\{} \\
\quad \texttt{"split": [(old\_key1, [new\_key1, new\_key2]), (old\_key2, [new\_key1, new\_key2])],} \\
\quad \texttt{"reparent": [(parent\_key1, [old\_child\_key1, old\_child\_key2]), (parent\_key2, [old\_child\_key1, old\_child\_key2])],} \\
\quad \texttt{"delete": [old\_key\_to\_delete1, old\_key\_to\_delete2],} \\
\quad \texttt{"merge": [(new\_key1, [old\_key\_to\_merge1, old\_key\_to\_merge2]), (new\_key2, [old\_key\_to\_merge1, old\_key\_to\_merge2])],} \\
\quad \texttt{"keep": [old\_key1, old\_key2]} \\
\texttt{\}} \texttt{</decision>} \\
If there is no operation for a certain task, do not display that item in the final output. \\

\#\# Current Layer Nodes \\
\{current\_keys\} \\

\#\# Leaf Nodes in the Current Layer \\
\{leaf\_keys\}
\end{promptbox}
\caption{Prompt of LLM-as-Decision-Maker (node refinement and pruning).}
\label{fig:dm_node_refine_prun}
\end{figure*}

\subsection{RealMix}
\label{sec:appendix_realmix}

\subsubsection{Seed Data}
\label{sec:appendix_realmix_seed_data}

The prompt for labeling the domain is shown in Fig.\ref{fig:annotation_domain}. 

The prompts for query tagging are shown in Fig.\ref{fig:annotation_query_tags_writing}, Fig.\ref{fig:annotation_query_tags_roleplay}, Fig.\ref{fig:annotation_query_tags_knowledge}, Fig.\ref{fig:annotation_query_tags_coding}, Fig.\ref{fig:annotation_query_tags_mathematics}, and Fig.\ref{fig:annotation_query_tags_reasoning}. 

The prompts for quality labeling are shown in Fig.\ref{fig:annotation_quality_writing}, Fig.\ref{fig:annotation_quality_roleplay}, Fig.\ref{fig:annotation_quality_knowledge}, Fig.\ref{fig:annotation_quality_coding}, Fig.\ref{fig:annotation_quality_mathematics}, and Fig.~\ref{fig:annotation_quality_reasoning}. 

The seed data is available in our dataset repository. 

\begin{figure*}[!h]
\centering
\begin{promptbox}[Prompt of Domain Annotation]{light_green}
\small
You are tasked with categorizing a given query into one of the following six categories:\\  
1. **roleplay**\\  
2. **coding**\\  
3. **mathematics**\\  
4. **reasoning**\\  
5. **knowledge**\\  
6. **writing**\\  
7. **other**\\

For each query provided, determine the most appropriate category and output the result in lowcase enclosed within <domain> and </domain> tags.\\

**Example**:\\  
Query: "How do I write a compelling essay introduction?"\\  
Output: <domain>roleplay</domain>\\

**Now, analyze the following query**:\\
<|begin\_of\_query|>\\
\{query\}\\
<|end\_of\_query|>
\end{promptbox}
\caption{Prompt of domain annotation. }
\label{fig:annotation_domain}
\end{figure*}

\begin{figure*}[!h]
\centering
\begin{promptbox}[Prompt of Query Tags Annotation (writing)]{blue_dist}
\small
You are tasked with categorizing a given query into multiple tags based on the following hierarchical classification system.  \\ 
 \\ 
\#\#\# Classification System: \\ 
\{taxonomy\}   \\ 

\#\#\# Rules for Tagging: \\ 
1. **Hierarchy Rule**: If a query matches both a parent node and its child nodes, include both in the tags. \\ 
2. **Multiple Matches**: If a query matches multiple nodes at the same level, include all matching nodes. \\ 
3. **No Match**: If a query does not match any nodes under the second-level main categories, assign the tag "Other" to that category. \\ 
4. **Output Format**: The final output must be enclosed within `<tags>` and `</tags>` tags, and the tags should be provided as a JSON object where the keys are the basis for classification and the values are lists of matching tags. \\ 
 \\ 
\#\#\# Example: \\ 
Query: "I need help brainstorming ideas for a fantasy story involving a hidden kingdom and magical creatures." \\ 
Output: <tags>\{\{"Creative Stages": ["Conceptualization Stage", "Brainstorming", "World Building", "Creative Exploration", "Content Conceptualization"], "Writing Domains": ["Literary Writing", "Fiction", "Fantasy", "Story Creation"], "Styles": ["Other"]\}\}</tags> \\ 
 \\ 
\#\#\# Now, analyze the following query:   \\ 
<|begin\_of\_query|>   \\ 
\{query\}   \\ 
<|end\_of\_query|>
\end{promptbox}
\caption{Prompt of query tags annotation (writing). }
\label{fig:annotation_query_tags_writing}
\end{figure*}

\begin{figure*}[!h]
\centering
\begin{promptbox}[Prompt of Query Tags Annotation (roleplay)]{blue_dist}
\small
You are tasked with categorizing a given query into multiple tags based on the following hierarchical classification system.  \\ 
 \\ 
\#\#\# Classification System: \\ 
\{taxonomy\} \\ 
 \\ 
\#\#\# Rules for Tagging: \\ 
1. **Hierarchy Rule**: If a query matches both a parent node and its child nodes, include both in the tags. \\ 
2. **Multiple Matches**: If a query matches multiple nodes at the same level, include all matching nodes. \\ 
3. **No Match**: If a query does not match any nodes under the second-level main categories, assign the tag "Other" to that category. \\ 
4. **Output Format**: The final output must be enclosed within `<tags>` and `</tags>` tags, and the tags should be provided as a JSON object where the keys are the basis for classification and the values are lists of matching tags. \\ 
 \\ 
\#\#\# Example: \\ 
Query: "Write a script for a medieval fantasy story involving a knight’s adventure and a magical encounter, with a humorous twist." \\ 
Output: <tags>\{\{"Theme Types": ["Fantasy", "Medieval", "Magic", "Comedy", "Humorous Comedy"], "Task Types": ["Creative", "Scriptwriting"], "Style Types": ["Humorous Style", "Light Humor", "Narrative"]\}\}</tags> \\ 
 \\ 
\#\#\# Now, analyze the following query:   \\ 
<|begin\_of\_query|>   \\ 
\{query\}   \\ 
<|end\_of\_query|>
\end{promptbox}
\caption{Prompt of query tags annotation (roleplay). }
\label{fig:annotation_query_tags_roleplay}
\end{figure*}

\begin{figure*}[!h]
\centering
\begin{promptbox}[Prompt of Query Tags Annotation (knowledge)]{blue_dist}
\small
You are tasked with categorizing a given query into multiple tags based on the following hierarchical classification system.  \\ 
 \\ 
\#\#\# Classification System: \\ 
\{taxonomy\} \\ 
 \\ 
\#\#\# Rules for Tagging: \\ 
1. **Hierarchy Rule**: If a query matches both a parent node and its child nodes, include both in the tags. \\ 
2. **Multiple Matches**: If a query matches multiple nodes at the same level, include all matching nodes. \\ 
3. **No Match**: If a query does not match any nodes under the second-level main categories, assign the tag "Other" to that category. \\ 
4. **Output Format**: The final output must be enclosed within `<tags>` and `</tags>` tags, and the tags should be provided as a JSON object where the keys are the basis for classification and the values are lists of matching tags. \\ 
 \\ 
\#\#\# Example: \\ 
Query: "How can I analyze data from a physics experiment on thermodynamics and present it effectively?" \\ 
Output: <tags>\{\{"Disciplinary Fields": ["Natural Sciences", "Physics", "Thermodynamics"], "Cognitive Levels": ["Applied Analysis"], "Task Types": ["Information Processing", "Data Analysis", "Content Production", "Content Generation"]\}\}</tags> \\ 
 \\ 
\#\#\# Now, analyze the following query:   \\ 
<|begin\_of\_query|>   \\ 
\{query\}   \\ 
<|end\_of\_query|>
\end{promptbox}
\caption{Prompt of query tags annotation (knowledge). }
\label{fig:annotation_query_tags_knowledge}
\end{figure*}

\begin{figure*}[!h]
\centering
\begin{promptbox}[Prompt of Query Tags Annotation (coding)]{blue_dist}
\small
You are tasked with categorizing a given query into multiple tags based on the following hierarchical classification system.  \\ 
 \\ 
\#\#\# Classification System: \\ 
\{taxonomy\} \\ 
 \\ 
\#\#\# Rules for Tagging: \\ 
1. **Hierarchy Rule**: If a query matches both a parent node and its child nodes, include both in the tags. \\ 
2. **Multiple Matches**: If a query matches multiple nodes at the same level, include all matching nodes. \\ 
3. **No Match**: If a query does not match any nodes under the second-level main categories, assign the tag "Other" to that category. \\ 
4. **Output Format**: The final output must be enclosed within `<tags>` and `</tags>` tags, and the tags should be provided as a JSON object where the keys are the basis for classification and the values are lists of matching tags. \\ 
 \\ 
\#\#\# Example: \\ 
Query: "How can I optimize a Python script that processes large datasets and visualizes the results?" \\ 
Output: <tags>\{\{"Task Types": ["Data Processing", "Data Analysis", "Data Visualization", "Debugging and Optimization", "Optimization"], "Technical Domains": ["Data Science", "Data Analysis Methods", "Data Visualization Tools"], "Programming Languages": ["General-purpose Languages", "Python"]\}\}</tags> \\ 
 \\ 
\#\#\# Now, analyze the following query:   \\ 
<|begin\_of\_query|>   \\ 
\{query\}   \\ 
<|end\_of\_query|>
\end{promptbox}
\caption{Prompt of query tags annotation (coding). }
\label{fig:annotation_query_tags_coding}
\end{figure*}

\begin{figure*}[!h]
\centering
\begin{promptbox}[Prompt of Query Tags Annotation (mathematics)]{blue_dist}
\small
You are tasked with categorizing a given query into multiple tags based on the following hierarchical classification system.  \\ 
 \\ 
\#\#\# Classification System: \\ 
\{taxonomy\} \\ 
 \\ 
\#\#\# Rules for Tagging: \\ 
1. **Hierarchy Rule**: If a query matches both a parent node and its child nodes, include both in the tags. \\ 
2. **Multiple Matches**: If a query matches multiple nodes at the same level, include all matching nodes. \\ 
3. **No Match**: If a query does not match any nodes under the second-level main categories, assign the tag "Other" to that category. \\ 
4. **Output Format**: The final output must be enclosed within `<tags>` and `</tags>` tags, and the tags should be provided as a JSON object where the keys are the basis for classification and the values are lists of matching tags. \\ 
 \\ 
\#\#\# Example: \\ 
Query: "How do I use calculus to model the growth of a population over time and graph the results?" \\ 
Output: <tags>\{\{"Mathematical Subfields": ["Analysis", "Calculus", "Applied Mathematics"], "Task Types": ["Problem Solving", "Modeling", "Visualization", "Function Graphing"]\}\}</tags> \\ 
 \\ 
\#\#\# Now, analyze the following query:   \\ 
<|begin\_of\_query|>   \\ 
\{query\}   \\ 
<|end\_of\_query|>
\end{promptbox}
\caption{Prompt of query tags annotation (mathematics). }
\label{fig:annotation_query_tags_mathematics}
\end{figure*}

\begin{figure*}[!h]
\centering
\begin{promptbox}[Prompt of Query Tags Annotation (reasoning)]{blue_dist}
\small
You are tasked with categorizing a given query into multiple tags based on the following hierarchical classification system.  \\ 
 \\ 
\#\#\# Classification System: \\ 
\{taxonomy\} \\ 
 \\ 
\#\#\# Rules for Tagging: \\ 
1. **Hierarchy Rule**: If a query matches both a parent node and its child nodes, include both in the tags. \\ 
2. **Multiple Matches**: If a query matches multiple nodes at the same level, include all matching nodes. \\ 
3. **No Match**: If a query does not match any nodes under the second-level main categories, assign the tag "Other" to that category. \\ 
4. **Output Format**: The final output must be enclosed within `<tags>` and `</tags>` tags, and the tags should be provided as a JSON object where the keys are the basis for classification and the values are lists of matching tags. \\ 
 \\ 
\#\#\# Example: \\ 
Query: "How can I determine the cause of an event based on multiple contributing factors?"   \\ 
Output: <tags>\{\{"Reasoning Methods": ["Multi-factor Attribution"], "Application Domains": ["Other"], "Thinking Modes": ["Analytical"], "Task Types": ["Reasoning"]\}\}</tags> \\ 
 \\ 
\#\#\# Now, analyze the following query:   \\ 
<|begin\_of\_query|>   \\ 
\{query\}   \\ 
<|end\_of\_query|>
\end{promptbox}
\caption{Prompt of query tags annotation (reasoning). }
\label{fig:annotation_query_tags_reasoning}
\end{figure*}

\begin{figure*}[!h]
\centering
\begin{promptbox}[Prompt of Quality Annotation (writing)]{yellow_dist}
\small
\#\# Task \\ 
Your task is to assess the quality of a given query based on the checklists outlined below. For each checklist, provide a clear explanation of your reasoning before assigning "Yes" or "No". If your answer is "Borderline", please answer "Yes" for this checklist.  \\ 
 \\ 
\#\# Checklists \\ 
Determine whether the provided question meets the following criteria. Return a Python array listing the numbers of all satisfied criteria: \\ 
 \\ 
1. **Clarity**: Is the question clear and well-defined? \\ 
2. **Completeness**: Does the question provide enough information for the LLM to answer the question? \\ 
3. **Complexity**: Does the question have enough depth and challenge beyond simple fact recall? \\ 
4. **Real-World Application**: Is this writing task something that would be proposed in the real world? \\ 
5. **Professionalism**: Does it require professional capabilities or professional knowledge? \\ 
6. **Originality:** Does the question encourage or require originality? \\ 
7. **User's Requirements**: Does the user have clear, detailed, or unique requests that need to be considered in the response? \\ 
 \\ 
For example, if the question meets Clarity, Completeness, and Real-World Application, return `[1, 2, 4]`. \\ 
 \\ 
\#\# Final Output \\ 
 \\ 
For each question provided, return a Python dictionary in the following format: \\ 
```python \\ 
Final Labels: \{\{"question\_quality": [1, 3, 4]\}\} \\ 
``` \\ 
 \\ 
\#\# The Query You Should Process \\ 
<|begin\_of\_query|> \\ 
\{query\} \\ 
<|end\_of\_query|>
\end{promptbox}
\caption{Prompt of quality annotation (writing). }
\label{fig:annotation_quality_writing}
\end{figure*}

\begin{figure*}[!h]
\centering
\begin{promptbox}[Prompt of Quality Annotation (roleplay)]{yellow_dist}
\small
\#\# Task \\ 
Your task is to assess the quality of a given query based on the checklists outlined below. For each checklist, provide a clear explanation of your reasoning before assigning "Yes" or "No". If your answer is "Borderline", please answer "Yes" for this checklist.  \\ 
 \\ 
\#\# Checklists \\ 
Determine whether the provided question meets the following criteria. Return a Python array listing the numbers of all satisfied criteria: \\ 
 \\ 
1. **Clarity**: Is the question clear and well-defined? \\ 
2. **Completeness**: Does the question provide enough information for the LLM to answer the question? \\ 
3. **Complexity**: Does it involve in-depth understanding of any role-playing content, such as the psychology, characterization, and world-building of characters? \\ 
4. **Real-World Application**: Is this role-playing task something that would be proposed in the real world? \\ 
5. **Interactivity**: Does the question encourage meaningful interactions between characters, rather than single character? \\ 
6. **Engagement**: Does the task motivate active participation and emotional involvement from the audience or participants? \\ 
7. **Creativity:** Does it have creativity and novelty, or does solving it require creativity? \\ 
 \\ 
For example, if the question meets Clarity, Completeness, and Real-World Application, return `[1, 2, 4]`. \\ 
 \\ 
\#\# Final Output \\ 
 \\ 
For each question provided, return a Python dictionary in the following format: \\ 
```python \\ 
Final Labels: \{\{"question\_quality": [1, 3, 4]\}\} \\ 
``` \\ 
 \\ 
\#\# The Query You Should Process \\ 
<|begin\_of\_query|> \\ 
\{query\} \\ 
<|end\_of\_query|>
\end{promptbox}
\caption{Prompt of quality annotation (roleplay). }
\label{fig:annotation_quality_roleplay}
\end{figure*}

\begin{figure*}[!h]
\centering
\begin{promptbox}[Prompt of Quality Annotation (knowledge)]{yellow_dist}
\small
\#\# Task \\ 
Your task is to assess the quality of a given query based on the checklists outlined below. For each checklist, provide a clear explanation of your reasoning before assigning "Yes" or "No". If your answer is "Borderline", please answer "Yes" for this checklist.  \\ 
 \\ 
\#\# Checklists \\ 
Determine whether the provided question meets the following criteria. Return a Python array listing the numbers of all satisfied criteria: \\ 
 \\ 
1. **Clarity**: Is the question clear and well-defined? \\ 
2. **Completeness**: Does the question provide enough information for the LLM to answer the question? \\ 
3. **Complexity**: Does the question have enough depth and challenge beyond simple fact recall? \\ 
4. **Real-World Application**: Is the question something that would be encountered in real-world? \\ 
5. **Depth of Knowledge**: Does the question require deep expertise in the subject instead of just memory? \\ 
6. **Cross-Disciplinary**: Does the question involve cross-disciplinary aspects? \\ 
7. **Open-Endedness.**: Does the question encourage open-ended responses rather than simple “yes” or “no” answers, promoting deeper thinking? \\ 
 \\ 
For example, if the question meets Clarity, Completeness, and Real-World Application, return `[1, 2, 4]`. \\ 
 \\ 
\#\# Final Output \\ 
 \\ 
For each question provided, return a Python dictionary in the following format: \\ 
```python \\ 
Final Labels: \{\{"question\_quality": [1, 3, 4]\}\} \\ 
``` \\ 
 \\ 
\#\# The Query You Should Process \\ 
<|begin\_of\_query|> \\ 
\{query\} \\ 
<|end\_of\_query|>
\end{promptbox}
\caption{Prompt of quality annotation (knowledge). }
\label{fig:annotation_quality_knowledge}
\end{figure*}

\begin{figure*}[!h]
\centering
\begin{promptbox}[Prompt of Quality Annotation (coding)]{yellow_dist}
\small
\#\# Task \\ 
Your task is to assess the quality of a given query based on the checklists outlined below. For each checklist, provide a clear explanation of your reasoning before assigning "Yes" or "No". If your answer is "Borderline", please answer "No" for this checklist.  \\ 
 \\ 
\#\# Checklists \\ 
Determine whether the provided question meets the following criteria. Return a Python array listing the numbers of all satisfied criteria: \\ 
 \\ 
1. **Clarity**: Is the question clear and well-defined? \\ 
2. **Completeness**: Does the question provide enough information for the LLM to answer the question? \\ 
3. **Complexity**: Does it involve multiple components, layers, or nuance? \\ 
4. **Real-World Application**: Is the question something that would be encountered in real-world development? \\ 
5. **Problem-Solving**: Does it require active problem-solving beyond simple and superficial script or fact recall? \\ 
6. **Domain-Specific Expertise**: Does the question require in-depth knowledge of at least one specific area of programming? \\ 
7. **Specified Requirements**: Does it specify particular requirements, such as execution time, space constraints, specific programming language, tools, packages, etc.? \\ 
 \\ 
For example, if the question meets Clarity, Completeness, and Real-World Application, return `[1, 2, 4]`. \\ 
 \\ 
\#\# Final Output \\ 
 \\ 
For each question provided, return a Python dictionary in the following format: \\ 
```python \\ 
Final Labels: \{\{"question\_quality": [1, 3, 4]\}\} \\ 
``` \\ 
 \\ 
\#\# The Query You Should Process \\ 
<|begin\_of\_query|> \\ 
\{query\} \\ 
<|end\_of\_query|>
\end{promptbox}
\caption{Prompt of quality annotation (coding). }
\label{fig:annotation_quality_coding}
\end{figure*}

\begin{figure*}[!h]
\centering
\begin{promptbox}[Prompt of Quality Annotation (mathematics)]{yellow_dist}
\small
\#\# Task \\ 
Your task is to assess the query of a given question based on the rules outlined below. For each rule, provide a clear explanation of your reasoning before assigning a label. \\ 
 \\ 
\#\# Rule \\ 
Determine whether the provided query meets the following criteria. Return a Python array listing the numbers of all satisfied criteria: \\ 
 \\ 
1. **Clarity**: Is the question clear and well-defined? \\ 
2. **Completeness**: Does the question provide enough information for the LLM to answer the question? \\ 
3. **Complexity**: Does it involve multiple steps, analysis, or reasoning instead of simple concept memorization and numerical calculation? \\ 
4. **Real-World Application**: Is the question something that would be encountered in real-world? \\ 
5. **Problem-Solving**: Does it test the ability to apply math in some scenarios? \\ 
6. **Rigorous Logic**: Does it involve content such as theorem derivation and formula understanding, which require rigorous logical abilities? \\ 
7. **Creativity:** Does it have creativity and novelty, or does solving it require creativity? \\ 
 \\ 
For example, if the question meets Clarity, Completeness, and Real-World Application, return `[1, 2, 4]`. \\ 
 \\ 
\#\# Final Output \\ 
 \\ 
For each question provided, return a Python dictionary in the following format: \\ 
```python \\ 
Final Labels: \{\{"question\_quality": [1, 3, 4]\}\} \\ 
``` \\ 
 \\ 
\#\# The Query You Should Process \\ 
<|begin\_of\_query|> \\ 
\{query\} \\ 
<|end\_of\_query|>
\end{promptbox}
\caption{Prompt of quality annotation (mathematics). }
\label{fig:annotation_quality_mathematics}
\end{figure*}

\begin{figure*}[!h]
\centering
\begin{promptbox}[Prompt of Quality Annotation (reasoning)]{yellow_dist}
\small
\#\# Task \\ 
Your task is to assess the query of a given question based on the rules outlined below. For each rule, provide a clear explanation of your reasoning before assigning a label. \\ 
 \\ 
\#\# Rule \\ 
Determine whether the provided query meets the following criteria. Return a Python array listing the numbers of all satisfied criteria: \\ 
 \\ 
1. **Clarity**: Is the question clear and well-defined? \\ 
2. **Completeness**: Does the question provide enough information for the LLM to answer the question? \\ 
3. **Complexity**: Does it involve multiple steps, analysis, or reasoning instead of simple concept memorization? \\ 
4. **Real-World Application**: Is the question something that would be encountered in real-world? \\ 
5. **Problem-Solving**: Does it require devising a solution or strategy?   \\ 
6. **Deep Thinking**: Does it require deep reasoning and consideration of multiple factors?   \\ 
7. **Novelty:** Does the question present a unique or unusual scenario that the LLM is unlikely to have encountered before? \\ 
 \\ 
For example, if the question meets Clarity, Completeness, and Real-World Application, return `[1, 2, 4]`. \\ 
 \\ 
\#\# Final Output \\ 
 \\ 
For each question provided, return a Python dictionary in the following format: \\ 
```python \\ 
Final Labels: \{\{"question\_quality": [1, 3, 4]\}\} \\ 
``` \\ 
 \\ 
\#\# The Query You Should Process \\ 
<|begin\_of\_query|> \\ 
\{query\} \\ 
<|end\_of\_query|>
\end{promptbox}
\caption{Prompt of quality annotation (reasoning). }
\label{fig:annotation_quality_reasoning}
\end{figure*}

\subsubsection{Synthesize New Queries}
\label{sec:appendix_realmix_syn_query}

The prompt of RealMix is shown in Fig.~\ref{fig:real_mix_prompt}

\begin{figure*}[!h]
\centering
\begin{promptbox}[Prompt of RealMix]{medium_purple}
\small
I will provide three real user questions and a reference question’s type tags. Your task is to create a new question in the \{domain\} domain with the same tags as reference question. To ensure the question feels authentic, you should utilize real-world details drawn from the three real user questions.  \\ 
 \\ 
\#\#\# You MUST follow these steps to generate the new question: \\ 
 \\ 
\#\#\#\# Task1: Real-world Details Selection: \\ 
 \\ 
Select a few real-world details (such as the real people, objects, scenes, settings, and any other details mentioned) from three real user questions that fit the given tags. If there are no suitable details, return [[I cannot generate a question based on the provided real user questions.]] and stop. \\ 
 \\ 
\#\#\#\# Task2: New Question Generation: \\ 
 \\ 
1. Although you should use details from real user questions, you must not mention the real user question in the new question. \\ 
2. The new question should be complex and challenging, requiring deep understanding and analysis of the subject. The length of the question should be at least as long as the reference question but should not be overly simplistic or repetitive. The question should be singular, not a multi-task question. \\ 
3. The new question must be **completely self-contained**, so that others can answer it without any additional information.  \\ 
4. Analyze how to create the new question with chosen real-world details and provided tags. While multiple tags are available, the newly generated question only needs to align with some of them, not all. Even if the original question already fits, generate a different version. \\ 
 \\ 
\#\#\# Output Format: \\ 
 \\ 
\texttt{[Anylysis]}: You should first complete the anylysis of task1 and task2 here.  \\ 
\texttt{[Question]}: Summarize the newly generated question in the following format: <new\_query>Insert the final new question here.</new\_query> \\ 
 \\ 
--- \\ 
 \\ 
<|begin\_of\_reference\_query|> \\ 
\{reference\_query\} \\ 
<|end\_of\_reference\_query|> \\ 
 \\ 
<|begin\_of\_reference\_query\_type\_tags|> \\ 
\{type\_tags\} \\ 
<|end\_of\_reference\_query\_type\_tags|> \\ 
 \\ 
<|begin\_of\_real\_user\_query\_1|> \\ 
\{query1\} \\ 
<|end\_of\_real\_user\_query\_1|> \\ 
 \\ 
<|begin\_of\_real\_user\_query\_2|> \\ 
\{query2\} \\ 
<|end\_of\_real\_user\_query\_2|> \\ 
 \\ 
<|begin\_of\_real\_user\_query\_3|> \\ 
\{query3\} \\ 
<|end\_of\_real\_user\_query\_3|>
\end{promptbox}
\caption{Prompt of RealMix. }
\label{fig:real_mix_prompt}
\end{figure*}

\subsection{Detailed Description of Visualization and Analysis Toolkits}
\label{sec:vis_and_analysis_tool}

\noindent\textbf{Visualization.} Two visualization tools are available: one organizes results by model, enabling side-by-side comparisons across multiple models; the other organizes results by query type, allowing users to examine model rankings under specific query tags. 

\noindent\textbf{Analysis.} We have implemented four analysis tools including Automatic Weakness Reporter, Automatic Weakness Analyzer, Failure Mode Explorer, and Automatic Difference Reporter. 




\noindent\textbf{(1) Automatic Weakness Reporter} detects capability imbalances by comparing a model’s ranking at a specific evaluation node with its overall ranking. If the local ranking differs from the global ranking by more than a predefined threshold (set to 5 in our experiments), the node is flagged as a weakness (negative deviation) or as a strength (positive deviation). This enables rapid localization of domains where the model exhibits pronounced strengths or weaknesses. For example, \emph{Gemma-3-4B-IT} shows its largest deviation in \emph{root.mathematics.Task Types.Visualization.Geometric Drawing}, where its overall ranking is 9 but its ranking in \emph{Geometric Drawing} drops to 16, revealing a substantial weakness in spatial and geometric visualization tasks.  

\noindent\textbf{(2) Automatic Weakness Analyzer} builds on the results of the Automatic Weakness Reporter and uses an advanced LLM (QwQ-32B in our study) to transform numerical deviations into human-readable summaries, grouping related deficiencies and strengths. This allows evaluators to understand the qualitative nature of the weaknesses detected. For \emph{Gemma-3-4B-IT}, for instance, the analyzer output includes:  
``Coding \& Tool Usage: Struggles with Swift, Batch, Lua programming languages; IDE configuration and test development.''  
Such summaries enable model developers to target specific technical capabilities for improvement.  

\noindent\textbf{(3) Failure Mode Explorer} investigates whether underperformance in a hierarchical category is comprehensive or localized. For any underperforming parent node, it calculates the standard deviation (STD) of rankings across its child nodes. If the STD is within the top 20\% of all nodes, the performance is considered \emph{unstable}, meaning that low performance may be caused by a few sub-capabilities. If the STD is in the bottom 20\%, the weakness is \emph{comprehensive} and spread across all sub-capabilities. For example, \emph{gemma} ranks 7 places worse in \emph{root.mathematics.Mathematical Subfields.Algebra} than its overall ranking, with a low STD of 0.4899. This indicates broad underperformance across Algebra, suggesting the need for holistic improvement rather than fixes in isolated subfields such as \emph{Linear Algebra}.  

\noindent\textbf{(4) Automatic Difference Reporter} enables model-to-model comparison by measuring differences in performance variability across evaluation nodes. The larger the difference in standard deviation between two models, the more divergent their capabilities are in that domain. This is useful for identifying areas of specialization or distinct model behavior. For instance, a comparison between \emph{claude-3.7-sonnet} and \emph{gpt-4o} reveals substantial differences in \emph{root.roleplay.Style Types.Experimental Style.Glitch Art}, \emph{root.roleplay.Theme Types.Emotional.Erotic}, and \emph{root.reasoning.Application Domains.Personal Development Guidance}, pointing to contrasting strengths and specialization patterns between the two models.  

Overall, this integrated framework transforms large-scale ranking data into actionable insights. By combining deviation detection, natural language summarization, hierarchical stability analysis, and comparative difference measurement, it supports interpretable evaluation and targeted optimization for LLM development.

\subsection{SCAN-D-V0}
\label{sec:appendix_evaluation_dataset}

\subsubsection{Query Quality: Human Evaluation}
\label{sec:appendix_evaluation_dataset_human_quality}

We conduct a comprehensive questionnaire study to evaluate the quality and human-likeness of automatically generated questions compared to original real-user questions. Five domain experts (four Master's holders and one PhD holder) participate in the questionnaire. The assessment protocol, shown in Fig.~\ref{fig:query_quality_and_reality_evaluation}, present participants with 20 question pairs (one generated and one original) in randomized order to prevent positional bias. 

\begin{figure*}[!h]
\centering
\begin{promptbox}[Query Quality and Reality Evaluation Questionnaire]{blue_dist}
\small
Dear Participant,

We are conducting a study on question generation for large language model. The attached Excel file contains 20 rows, each presenting two questions generated by different strategies. Please:

\begin{enumerate}
    \item Assess the \textbf{quality} of both questions (poor quality indicators include: oversimplification, incompleteness, or unclear phrasing). 
    \item Evaluate whether each question appears to be \textbf{human-authored}. 
\end{enumerate}

Rating Scheme:
\begin{itemize}
    \item For quality/authenticity: 
    \begin{itemize}
        \item 1 = Question 1 superior
        \item 2 = Question 2 superior
        \item 3 = Both inadequate
        \item 4 = Both excellent
    \end{itemize}
\end{itemize}
\end{promptbox}
\caption{Questionnaire protocol for comparative evaluation.}
\label{fig:query_quality_and_reality_evaluation}
\end{figure*}

Tab.~\ref{tab:query_quality_and_reality_evaluation_results} presents the detailed evaluation outcomes across all raters. Our generated questions demonstrate superior quality in 49\% of cases compared to just 9\% where original questions are preferred, with 42\% of cases rated as ties. For authenticity assessment, generated questions are preferred in 27\% of cases versus 21\% for originals, with a majority (52\%) considered equally authentic. Notably, 37 out of 100 cases are rated as having both excellent quality, while 47 cases are judged equally authentic. 

\begin{table*}[!h]
\centering
\resizebox{\textwidth}{!}{
\begin{tabular}{lcccccc}
\toprule
\textbf{Rating} & \textbf{Rater 1} & \textbf{Rater 2} & \textbf{Rater 3} & \textbf{Rater 4} & \textbf{Rater 5} & \textbf{Total} \\
& Quality/Reality & Quality/Reality & Quality/Reality & Quality/Reality & Quality/Reality & Quality/Reality \\
\midrule
1 (Generated better) & 11/5 & 15/8 & 10/3 & 7/7 & 6/4 & 49/27 \\
2 (Original better) & 0/4 & 4/10 & 4/7 & 0/0 & 1/0 & 9/21 \\
3 (Both inadequate) & 1/1 & 0/0 & 2/3 & 2/1 & 0/0 & 5/5 \\
4 (Both excellent) & 8/10 & 1/2 & 4/7 & 11/12 & 13/16 & 37/47 \\
\bottomrule
\end{tabular}}
\caption{Query quality and reality evaluation results by human rater.}
\label{tab:query_quality_and_reality_evaluation_results}
\end{table*}

\subsubsection{Query Quality: Data Contamination}
\label{sec:appendix_evaluation_dataset_data_contamination}

Following Auto-Arena~\cite{zhao2024auto}, we employ two methods to detect test data contamination. The first is the string match metric, as used in GPT-4~\cite{achiam2023gpt}, which flags a data point as contaminated if any of three randomly sampled 50-character substrings (or the entire string, if shorter) from the evaluation sample appears as a substring in the training set. The second method, inspired by Platypus~\cite{lee2023platypus}, uses sentence embedding similarity: a question is considered contaminated if its BERT-Large~\cite{devlin2019bert} embedding has a cosine similarity above 0.8 with any training item, enabling detection of paraphrased overlaps. 

Applying these methods, we find a contamination rate of 1.41\% using the substring match metric, and a substantially lower rate of 0.21\% using the embedding similarity metric. These results indicate that our approach is largely robust to data leakage and that potential contamination is minimal.

\subsubsection{Examples of SCAN-D-V0}
\label{sec:example_dataset}

We present several samples of SCAN-D-V0 in Tab.~\ref{tab:dataset_samples}. 

\newcolumntype{L}{>{\raggedright\arraybackslash}p{3cm}} 
\newcolumntype{R}{>{\raggedright\arraybackslash}p{9cm}} 
\begin{table*}[!h]
\centering
\begin{tabular}{>{\raggedright\arraybackslash}p{1.5cm} >{\raggedright\arraybackslash}p{11.5cm}}
\toprule
\textbf{Domain} & \textbf{Sample} \\
\midrule
Writing & Translate and adapt the following mission dialogue into French, ensuring it resonates with a native French speaker. The dialogue should maintain its dramatic tone while incorporating cultural nuances and idiomatic expressions suitable for a fantasy RPG setting. Additionally, optimize the sentence structure to enhance flow and engagement. "Listen carefully, adventurer. The ancient temple lies hidden beyond the Whispering Woods. Within its walls, you will find the Earth Crystal, a source of immense power. Use it wisely, for it can either restore balance or bring about destruction. Be wary of the Guardians; they will test your worth. Prove yourself, and the crystal shall be yours." \\
\hline
Roleplay & Please pretend to be my deceased grandfather, who was a gentle and wise artist from the small coastal town of Lestupsk. He used to live in a house near the old fortress on the hill and would often take me to ride the drowsy trams that clanked through our town. He taught me how to draw and always shared stories about his friend, Vanya, another local artist. I miss him so much, especially when I am trying to fall asleep. We begin now. Hello grandpa, I miss you so much! I am feeling very tired and am having trouble falling asleep. Can you please tell me one of those special stories about Vanya and the old trams, and maybe show me how to sketch the view of the fortress from our window? Your stories always made me feel better and helped me sleep. \\
\hline
Knowledge & Analyze how the infrared absorption spectrum of a greenhouse gas (e.g., CO2 or methane) impacts the heat transfer efficiency in a thermodynamic system such as a gas turbine or a heat exchanger. Propose strategies to counteract any detrimental effects on system performance caused by such radiative properties while maintaining environmental compliance. \\
\hline
Coding & In Java, which method is commonly used to remove quotes from the beginning and end of a string? a) `trim()` b) `replaceAll("\"", "")` c) `substring(1, str.length() - 1)` d) `strip()` e) None of the above \\
\hline
Mathematics & Let \( F = A'B'C' + A'B'C + A'BC' + AB'C' + ABC' \). Using Boolean algebraic laws, simplify the expression \( F \) to its minimal form. Identify which laws or manipulations you use at each step to arrive at the solution. \\
\hline
Reasoning & A teacher places a book inside a drawer of her desk in the classroom. She then locks the drawer and places the key inside a box on her shelf. After school, the teacher moves the box to the staff room and leaves it on a table. Later, a janitor takes the box from the staff room and places it inside a cabinet in the storage room. Where is the key, and where is the book? \\
\bottomrule
\end{tabular}
\caption{Examples from different domains in SCAN-D-V0. }
\label{tab:dataset_samples}
\end{table*}

\subsubsection{SCAN-HPD}
\label{sec:appendix_evaluation_dataset_hpd}

To validate our evaluation methods, we require a robust human preference dataset. We construct this dataset from seed data, which is well-suited for our purposes as it contains two responses and a corresponding human preference label for each query. We begin by randomly sampling 650 queries rated 6 or 7 from the seed data, followed by manual quality control, yielding a final set of 636 high-quality queries.

\begin{figure*}[!h]
\centering
    \includegraphics[width=\textwidth]{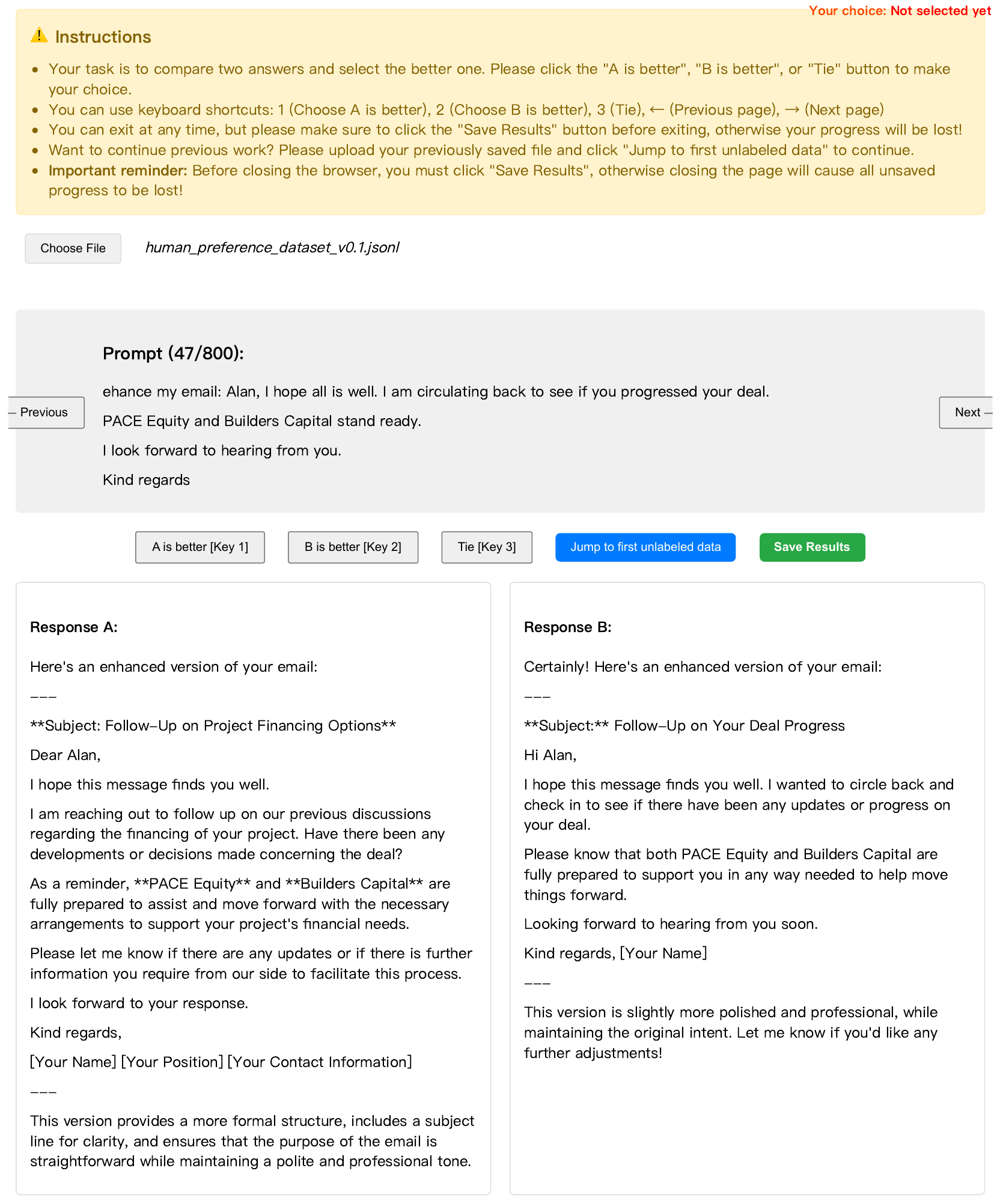}
\caption{The annotation interface used for labeling the human preference dataset.}
\label{fig:evaluation_dataset_hpd_html}
\end{figure*}

Although the seed data includes human preference labels, prior work~\cite{zheng2023judging} has raised concerns about their reliability due to variability in human judgment. To enhance label accuracy, we enlisted two additional graduate-level annotators and determined the final preference label via majority voting, thereby mitigating individual annotator bias. Our annotation interface is illustrated in Fig.~\ref{fig:evaluation_dataset_hpd_html}. 

Since our questions are all compliant and commonly seen in real-world scenarios, they do not pose significant potential participant risks. Additionally, we pay 300 yuan per day for each participant, and we ultimately spent 1,200 yuan. 

\begin{figure*}[!h]
\centering
    \includegraphics[width=0.5\textwidth]{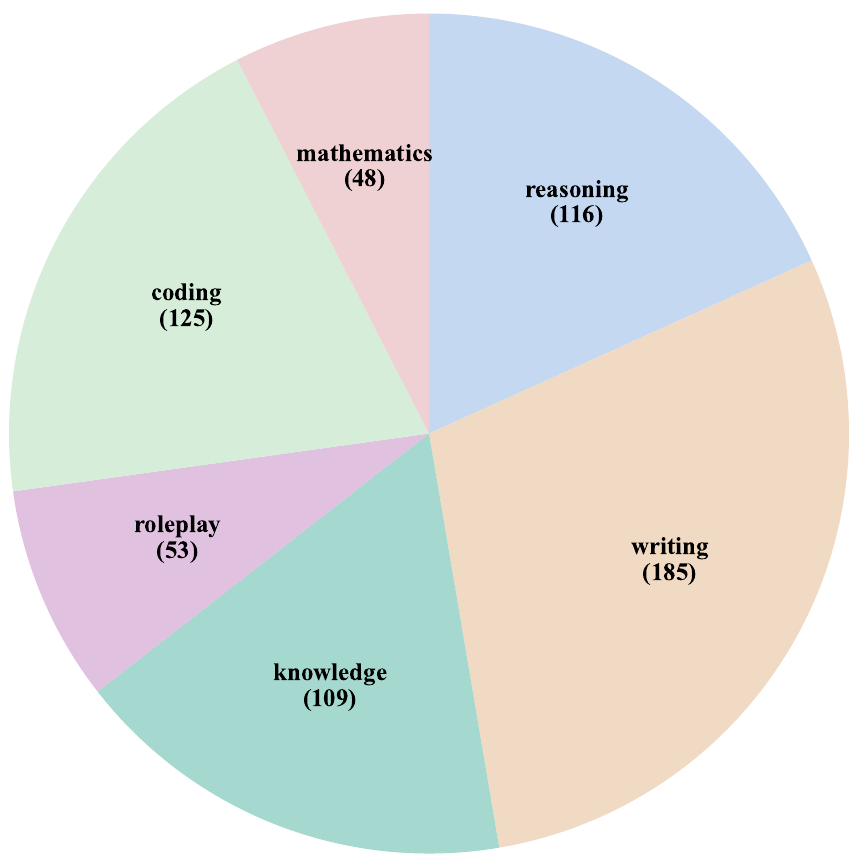}
\caption{Domain distribution of SCAN-HPD.}
\label{fig:domain_distribution_hpd}
\end{figure*}

The distribution of SCAN-HPD is shown in Fig.~\ref{fig:domain_distribution_hpd}. Similar to SCAN-D-V0, these queries span six domains: writing, roleplay, knowledge, coding, mathematics, and reasoning. 

\section{Evaluation Methods}
\label{sec:supplementary_material_of_eval_method}

\begin{figure*}[!h]
\centering
\begin{promptbox}[Example (1) of Pairwise Evaluation]{blue_dist}
\scriptsize
(prompt and thinking process) ...
\begin{verbatim}
[[A]]
\end{verbatim}
**Explanation:** \\ 
- **Correctness:** Assistant A's answer directly addresses the user's requirement to track the *number of open tabs* on the same domain using `localStorage`. It provides a straightforward `TabTracker` class with methods to increment/decrement the count and store it in `localStorage`. While it lacks automatic triggering of decrement on tab closure (relying on manual calls in the example), it correctly structures the core logic for counting tabs. The problem does not explicitly require automatic event handling, so the code is technically correct for the stated goal. \\ 
 \\ 
- **Helpfulness:** The code is concise and focused on the core requirement. The explanation clearly describes how each method works and how to use the class. The example usage demonstrates the basic workflow, making it easy to integrate into a project. \\ 
 \\ 
- **Assistant B's Issues:** While B's approach is more detailed (tracking individual tabs with IDs and URLs), it diverges from the user's request to track *open tabs as a count*. B's implementation requires managing individual tabs and resolving complex edge cases (e.g., identifying which tab closed), which complicates the task. Additionally, B's `\_handleTabClose` method admits it cannot reliably remove tabs, making the solution incomplete for the stated problem. \\ 
 \\ 
- **Key Flaw in B's Answer:** The user asked for a *count* of open tabs, not detailed tracking of individual tabs. B's answer overcomplicates the solution by focusing on tab metadata rather than the numeric count. \\ 
 \\ 
**Conclusion:** Assistant A’s answer is better because it aligns directly with the user’s requirement, provides a functional core, and is simpler and more relevant. Assistant B’s answer, while thorough, solves a different problem (tracking individual tabs) and introduces unnecessary complexity.
\end{promptbox}
\caption{Example (1) of Pairwise Evaluation.}
\label{fig:example_pairwise_evaluation_1}
\end{figure*}

\begin{figure*}[!h]
\centering
\begin{promptbox}[Example (2) of Pairwise Evaluation]{blue_dist}
\scriptsize
(prompt and thinking process) ...
\begin{verbatim}
[[B]]
\end{verbatim}
**Explanation:**   \\ 
Both assistants provided valid HTML documents with step-by-step instructions for jump-starting a car. However, **Assistant B's answer is better** for the following reasons:   \\ 
 \\ 
1. **Self-Contained Diagram:**   \\ 
   - **Assistant B** uses an embedded SVG diagram directly in the HTML, fulfilling the user's requirement for an "all in one document." The SVG illustrates the connection process without relying on an external image file.   \\ 
   - **Assistant A** references an external image (`car\_battery\_jumpstart.png`), which requires the user to provide or replace the image path. This makes the document incomplete as-is.   \\ 
 \\ 
2. **Step-by-Step Clarity:**   \\ 
   - **Assistant B** breaks down the steps into a more granular 8-step process, explicitly separating each cable connection into distinct steps (e.g., connecting each end of the red and black cables separately). This enhances clarity for users unfamiliar with the process.   \\ 
   - **Assistant A** combines some steps (e.g., all cable connections under step 3 with sub-steps), which is concise but slightly less explicit.   \\ 
 \\ 
3. **Safety and Correctness:**   \\ 
   - Both answers include correct procedural steps (e.g., grounding the negative cable on the dead car’s metal surface instead of the terminal).   \\ 
   - **Assistant B** adds a final step (step 8) explicitly detailing the order to remove cables, which is important for safety.   \\ 
 \\ 
4. **HTML Structure:**   \\ 
   - **Assistant B’s** use of SVG ensures the diagram is part of the HTML document, whereas **Assistant A’s** reliance on an external image breaks the "all in one" requirement.   \\ 
 \\ 
**Verdict:** While both answers are correct, **Assistant B’s** self-contained implementation of the diagram and clearer step breakdown make it the superior choice. 
\end{promptbox}
\caption{Example (2) of Pairwise Evaluation.}
\label{fig:example_pairwise_evaluation_2}
\end{figure*}

\end{document}